\documentclass[runningheads]{llncs}

\usepackage{eccv}

\usepackage{eccvabbrv}

\usepackage{graphicx}
\usepackage{booktabs}
\usepackage{orcidlink}

\usepackage{multirow}
\usepackage{colortbl} 
\usepackage{amssymb}

\usepackage[accsupp]{axessibility}  
\usepackage{bigdelim} 

\usepackage{hyperref}

\begin{document}

\title{ENet-GP: Unified Document Image Restoration} 


\author{Sujal Burad\inst{1} \and
Aakanksha\inst{1} \and
A. N. Rajagopalan\inst{1} \and
Sumit Shekar\inst{2}
}

\authorrunning{S.Burad et al.}

\institute{Indian Institute of Technology, Madras\\
\email{sujalburad99@gmail.com, ee18d405@smail.iitm.ac.in, raju@ee.iitm.ac.in} \and
Adobe, India\\
\email{sushekha@adobe.com}}

\maketitle

\begin{abstract}
Reliable document digitization in uncontrolled capture settings is challenging because real images exhibit multiple interacting degradations rather than a single isolated distortion. Documents thus captured are affected simultaneously by geometric distortions, like page warping, as well as photometric degradations such as non-uniform illumination, and blurring. However, most existing approaches address these factors independently and are evaluated on benchmarks containing only one distortion type, limiting their real-world applicability. We introduce GutenDoc, a large-scale dataset of high-resolution dense-text documents with physically grounded compound degradations. Using physics-based rendering, our dataset jointly models geometric warping and diverse photometric effects, enabling systematic evaluation under realistic capture conditions. We further propose a unified restoration framework that jointly corrects geometric and photometric distortions within a single-network and single-training setup, without the need for degradation-specific retraining or sequential inference passes. Extensive experiments show that our method remains competitive on established single-distortion benchmarks while substantially improving robustness under compound degradations, providing a practical solution for real-world document digitization.

  \keywords{Document Restoration \and Unified framework \and Synthetic Data}
\end{abstract}

\begin{figure}[t]
  \centering
  \includegraphics[width=\columnwidth]{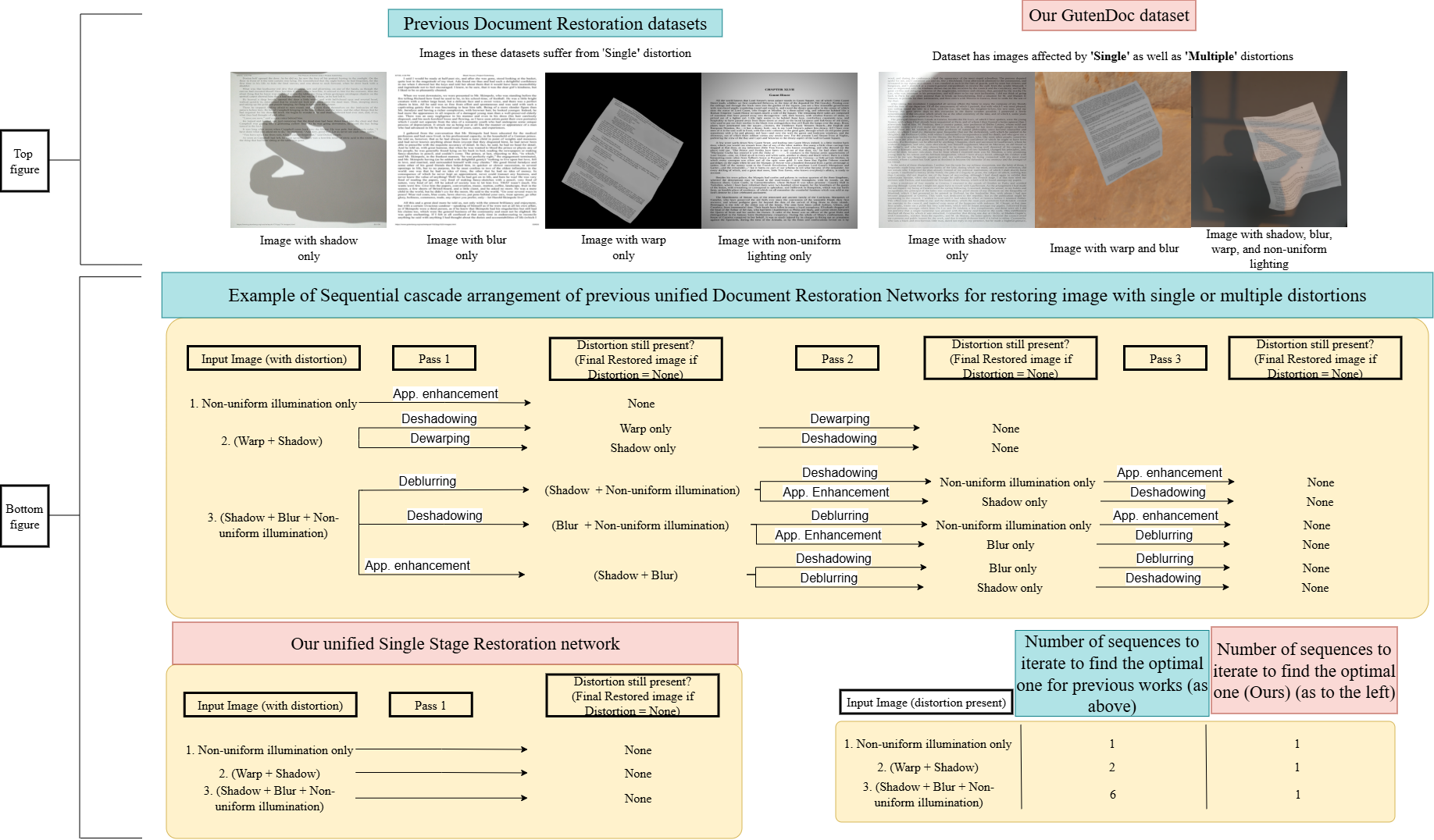}
  \caption{Top figure: Degradation comparison between previous Document Restoration datasets and our proposed document dataset, GutenDoc. 
  Bottom figure: Unlike previous cascaded methods that require sequential ordering and multiple passes with iterative evaluation to find the optimal ordering, our proposed unified network, ENet-GP restores any combination of document degradations in a single-pass using a single-network-single-training setup paradigm.
}
  \label{fig:overall_contri}
\end{figure}

\section{Introduction}
\label{sec:intro}

In real-world settings, document images captured by handheld devices often lack the uniform illumination and geometric alignment found in dedicated scanners. Document images captured in such imaging setups frequently suffer from different kinds of deviations making it difficult to use such documents, as is, in their digitized format. Such imperfections can be broadly be categorized into geometric distortions (e.g., perspective errors, warping, folds, etc) and photometric distortions (e.g., illumination non-uniformity, shadows, blur, etc), with each type introducing its own unique challenges for restoration. The presence of these degradations significantly impairs the performance of downstream document image processing tasks such as optical character recognition (OCR), layout analysis, text segmentation, and information extraction~\cite{kim2022ocrfreedocumentunderstandingtransformer, blecher2023nougat, davis2022endtoenddocumentrecognitionunderstanding, 4376991}.

To address these restoration challenges, numerous approaches have been introduced; however, most methods treat these distortions in isolation. Specifically, geometric distortion is addressed by \cite{das2019dewarpnet, zhang2025dvdunleashinggenerativeparadigm, Verhoeven_2023}, while photometric distortion is handled by \cite{feng2021docmae, 9156786, tensmeyer2017documentimagebinarizationfully, wang2022udocganunpaireddocumentillumination}. Although these models exhibit good performance on their respective tasks, they lack the ability to address multiple distortions in a single-network setup, thus requiring multiple degradation specific model configurations to restore for more than one distortion type.

DocRes \cite{zhang2024docresgeneralistmodelunifying}, on the other hand, unifies five restoration tasks: dewarping, deshadowing, appearance enhancement, deblurring, and binarization, via a prompting mechanism which guides task-specific execution and enhances performance via prior features extracted from the input image. UniDocDiff \cite{zhao2025unidocdiffunifieddocumentrestoration} adopts a diffusion-based framework with learnable task prompts and selective adaptability of the features relevant to each task. Despite these advances, these unified models still struggle with restoring images having simultaneous geometric and photometric distortions, especially when multiple distortions are entangled, in a single pass. They need to be arranged in a sequential cascade with multiple passes to attend to images with multiple distortions, as shown in the bottom of Figure \ref{fig:overall_contri}. Additionally, constrained priors limit the models' flexibility in diverse, uncertain real-world scenarios. \cite{10447446} proposes joint rectification but its evaluation is restricted to the geometrically-focused DocUNet benchmark \cite{Ma_2018_CVPR}. The model's versatility and robustness on diverse document images featuring complex entangled distortions has not been studied.

This underlines the need for a high-resolution dataset that contains dense text document images, to stress-test the restoration pipeline for downstream OCR capability, against simultaneous geometric and photometric distortions \cite{das2019dewarpnet, zhang2025docaligner, 9897217, Zhang_2023_CVPR, das2020intrinsicdecompositiondocumentimages, s20236929, 8583809, feng2022geometric}. While geometric distortions (e.g. warping, curling, and perspective deformations) require global spatial transformations and rectification-based solutions, photometric distortions (e.g., shadows, blur, and non-uniform illumination) require illumination correction techniques. These degradations are fundamentally coupled: shadows, illumination and blur are influenced by local geometric deformations (folds, warps, and curves) and global document structure. Furthermore, high-frequency text details are critically dependent on local geometric variations, creating an intricate photometric-geometric interplay. 

To advance research on document degradation, we introduce GutenDoc, a semi-realistic document dataset featuring high-resolution dense text images with entangled geometric degradation, consisting of warps, and photometric degradation, consisting of shadows, blur, and non-uniform illumination. Using physics-based rendering for generation, we entangle these distortions at varying severities to simulate complex real-word degradation, providing a controlled yet realistic benchmark for evaluating unified restoration models and studying the interplay among distortion types. This dataset will be publicly released. Subsequently, to address both Geometric and Photometric degradations simultaneously, we propose a unified architecture, ENet-GP (so named due to its $`$E$'$ shape), employing a common encoder that jointly learns Geometric and Photometric features. Two separate decoders operate on this common representation: one for geometric rectification (dewarping) and another for photometric correction (deshadowing, deblurring, and non-uniform illumination removal). Combining these decoder outputs achieves single stage document restoration --- a unified restoration approach largely unexplored despite extensive studies on isolated distortions~\cite{9879292,10447446, mao2016imagerestorationusingdeep, ruder2017overviewmultitasklearningdeep, 9706885}. Figure \ref{fig:overall_contri} depicts our broad framework and the proposed dataset. We also evaluate ENet-GP on existing benchmarks to demonstrate its robustness and generalization.

In summary, our main contributions are as follows: 
\begin{itemize}
    \item  We propose GutenDoc, a semi-realistically generated high-resolution dense text document image dataset simulating combined geometric and photometric real-world document degradations.

    \item We present a unified document restoration architecture, ENet-GP, to simultaneously address both geometric and photometric distortions, within a single-network-single-training setup paradigm, eliminating the need for multi-stage or task-specific models.
    
\end{itemize}

\section{Related Works}
\subsection{Document Image Restoration (DIR) Datasets and Architectures}
Geometric restoration, which primarily involves dewarping, maps spatially deformed documents (e.g., warped, folded) to their flat versions \cite{das2019dewarpnet, Ma_2018_CVPR, SagnikKeICCV2019, feng2021docmae, jiang2022revisiting, Li_2023_ICCV, zhao2025unidocdiffunifieddocumentrestoration, Hertlein_2025_WACV, zhang2024docresgeneralistmodelunifying, Verhoeven_2023, zhang2025dvdunleashinggenerativeparadigm, hertlein2025docmatcher, feng2022docscannerrobustdocumentimage}. Photometric restoration addresses visual legibility issues caused by shadows, blur, and color artifacts through tasks like deshadowing \cite{wang2022udocganunpaireddocumentillumination, 9156786, Zhang_2023_CVPR, hertlein2019scannet}, appearance enhancement \cite{Hertlein_ICCVW2023, das2020intrinsicdecompositiondocumentimages, feng2022doctrdocumentimagetransformer}, binarization \cite{calvo2019selectional, tensmeyer2017documentimagebinarizationfully, zhang2023docbinformer, Rezanezhad_2024_Hybrid}, and deblurring \cite{zamir2022restormer, souibgui2020docentr, zhang2024docresgeneralistmodelunifying, yang2023docdiffdocumentenhancementresidual}.

To address the scarcity of ground-truth data, researchers have introduced numerous datasets. Table \ref{tab:related_works_datasets} summarizes prominent synthetic and real-world datasets utilized for training and benchmarking. Despite the breadth of existing resources, current datasets exhibit significant limitations. First, the majority are predominantly low-resolution and lack densely populated text regions, degrading performance in fine-grained downstream tasks such as OCR, text segmentation, and information extraction. Second, existing protocols largely treat degradations in isolation, assuming a document suffers from only a single type of distortion. This fails to reflect real-world acquisition scenarios where documents are frequently subjected to concurrent geometric (e.g., folds, warping, perspective distortions) and photometric (e.g., shadows, blur) anomalies.

\begin{table*}[!t]
\centering
\caption{Summary of existing DIR datasets.}
\label{tab:related_works_datasets}
\resizebox{\textwidth}{!}{%
\begin{tabular}{p{2.5cm} p{6.8cm} @{\hspace{0.5cm}} p{6.8cm}}
\hline
\textbf{Task} & \textbf{Real-World Datasets (Size)} & \textbf{Synthetic Datasets (Size)} \\
\hline
\textbf{Dewarping} & DocUnet (130) \cite{Ma_2018_CVPR}, DIW(5k)\cite{10.1145/3528233.3530756}, DIR300 (300) \cite{feng2022geometricrepresentationlearningdocument}, Inv3DReal (360) \cite{Hertlein2023}, DocReal(200)\cite{Yu_2024_WACV}, WarpDoc-R(840)\cite{10.1145/3664647.3681548}, UDIR (195) \cite{10374269}, Book100 (100) \cite{liu2026booknetbookimagerectification}, WarpDoc (1k) \cite{9880145}, UVDoc (50) \cite{Verhoeven_2023} & Doc3D (100k) \cite{das2019dewarpnet}, UVDoc (pseudo-realistic)(20k) \cite{Verhoeven_2023}, Inv3d (25k) \cite{Hertlein2023}, DICP (30k) \cite{xie2022documentdewarpingcontrolpoints}, \cite{10.1145/3627818} (90k), Book3D (56k) \cite{liu2026booknetbookimagerectification}, AlignSynth \cite{zhang2025docaligner} \\
\hline
\textbf{Deshadowing} & RDD (4.9k) \cite{Zhang_2023_CVPR}, Kligler (300) \cite{8578350}, Jung (87) \cite{jung2019waterfillingefficientalgorithmdigitized}, OSR (237) \cite{s20236929}, WEZUT (176) \cite{article3} & FSDSRD (14.2k) \cite{9897217}, SynDocDS (50k) \cite{article2}, SD7k \cite{li2024highresolutiondocumentshadowremoval} \\
\hline
\textbf{Appearance} & DocAligner (130) \cite{zhang2025docaligner}, RealDAE (600) \cite{10268585} & Doc3DShade (90k) \cite{das2020intrinsicdecompositiondocumentimages}, DocProj (2.4k) \cite{10.1145/3355089.3356563} \\
\hline
\textbf{Binarization} & (H)-DIBCO\cite{8583809, 5277767, 6981120, 5693650, 6065249, 6424498, 6628857, 7814134, 8270159, 8978205}, Synchromedia(240)\cite{7333947}, Persian Heritage(15)\cite{Ayatollahi_2013}, Bickley Diary (7) \cite{10.1145/1816123.1816161} & Noisy Office (216) \cite{10.5555/1768409.1768429} \\
\hline
\textbf{Deblurring} & \multicolumn{2}{l}{TDD \cite{inproceedings} (used for both training and benchmarking)} \\
\hline
\end{tabular}%
}
\end{table*}

\begin{table*}[!b]
\caption{Overview of DIR architectures.}
\label{tab:architectures}
\resizebox{\textwidth}{!}{%
\begin{tabular}{p{2.5cm} p{4cm} @{\hspace{0.5cm}} p{5cm} @{\hspace{0.5cm}} p{6cm}}
\hline
\textbf{Category} & \textbf{Representative Methods} & \textbf{Core Mechanisms} & \textbf{Primary Limitations} \\
\hline
\textbf{Geometric Networks} & DocUNet \cite{Ma_2018_CVPR}, DewarpNet \cite{das2019dewarpnet}, DocTr \cite{feng2022doctrdocumentimagetransformer}, DvD \cite{zhang2025dvdunleashinggenerativeparadigm} & U-Net, 3D coordinate prediction, Transformers, Diffusion models & Treat shape correction in isolation; fail to address accompanying photometric distortions. \\
\hline
\textbf{Photometric Networks} & Skip-connections \cite{he2015deepresiduallearningimage, mao2016imagerestorationusingdeep, 8546199}, General \cite{abuolaim2020defocusdeblurringusingdualpixel, cho2021rethinkingcoarsetofineapproachsingle, zamir2021multistageprogressiveimagerestoration}, SwinIR \cite{liang2021swinirimagerestorationusing}, Restormer \cite{zamir2022restormer}, Uformer \cite{wang2021uformergeneralushapedtransformer}, DocEnTr \cite{souibgui2020docentr}, DocDeshadower \cite{zhou2024docdeshadowerfrequencyawaretransformerdocument}, BGSNet \cite{Zhang_2023_CVPR}, DocDiff \cite{yang2023docdiffdocumentenhancementresidual} & Residual learning, Vision Transformers (Self-attention, Frequency-aware), Diffusion models & Assume a spatially aligned, flat surface; struggle with geometry-induced lighting variations. \\
\hline
\textbf{Unified Solvers} & ProRes \cite{ma2023proresexploringdegradationawarevisual}, DocRes \cite{zhang2024docresgeneralistmodelunifying}, UniDocDiff \cite{zhao2025unidocdiffunifieddocumentrestoration}, Tang et al. \cite{10447446}, DocNLC \cite{wang2024docnlc} & Learnable/Dynamic prompting, Diffusion, Dual-stream networks, Contrastive learning & Struggle to synergistically model the intrinsic correlation between 3D deformations and photometric distortions. \\
\hline
\end{tabular}%
}
\end{table*}

The document restoration architectures generally follow three distinct paradigms: geometric correction, photometric enhancement, and unified solvers. Table \ref{tab:architectures} summarizes these approaches, their representative methods, and their inherent limitations. 

\begin{figure*}[ht]
    \centering
    
    \begin{minipage}{0.5\textwidth}
        \centering
        \includegraphics[width=\textwidth]{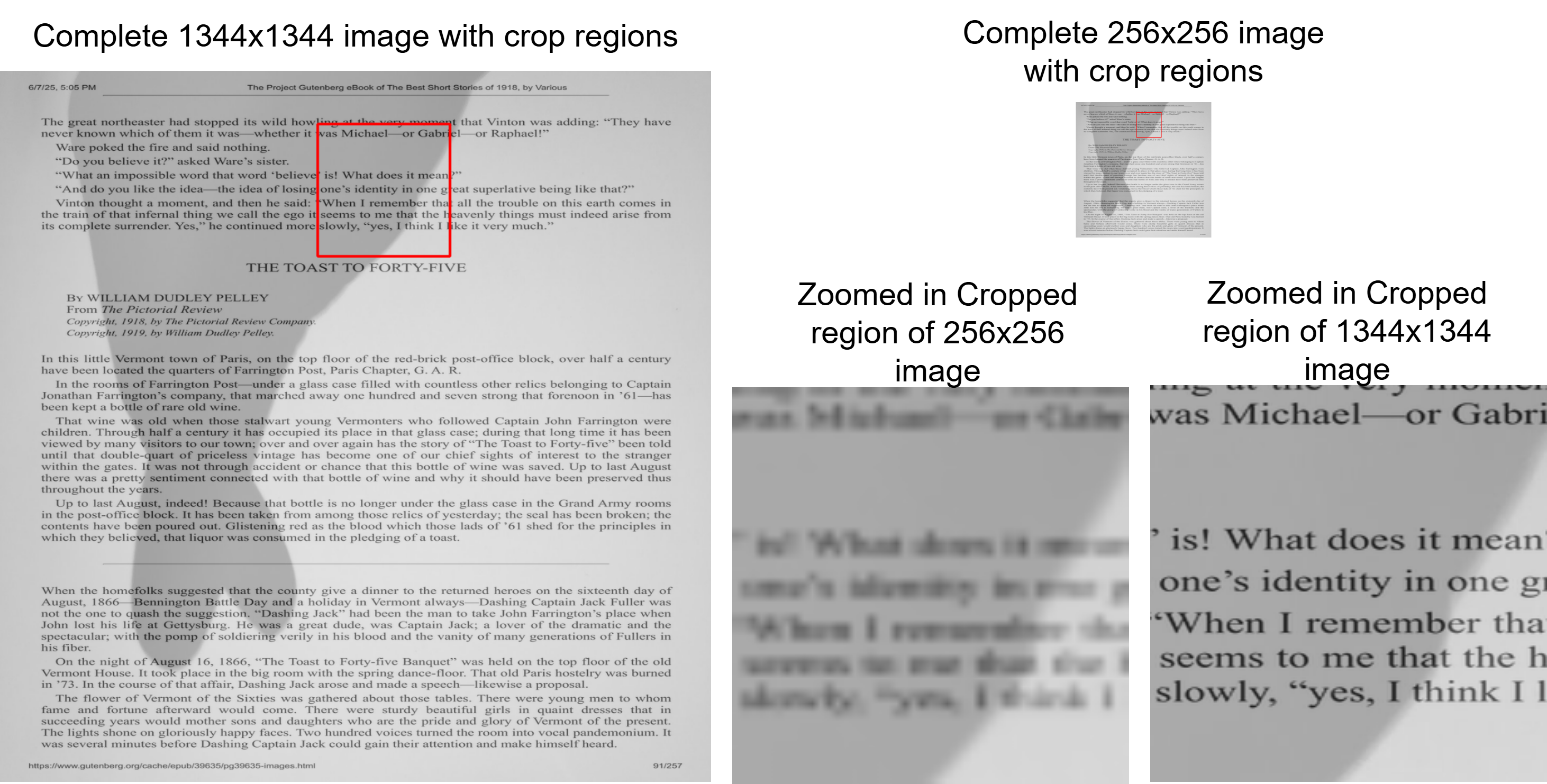}
        \caption{Identical shadow distortion comparison across resolutions. Zoomed crops reveal that the $256 \times 256$ image appears to suffer from additional blur, whereas the $1344 \times 1344$ version maintains edge clarity.}
        \label{1344vs256.drawio}
    \end{minipage}
    \hfill 
    \begin{minipage}{0.48\textwidth}
        \centering
        \captionof{table}{Distortion parameterization in GutenDoc. The dataset uses controlled sampling of geometric and photometric effects with predefined probability distributions to enable systematic evaluation.}
        \label{tab:distortion_params}        
        \renewcommand{\arraystretch}{1.8}
        \setlength{\tabcolsep}{2pt}
        \small
        \resizebox{\textwidth}{!}{
            \begin{tabular}{llccc}
            \toprule
            \textbf{Category} & \textbf{Distortion Type} & \textbf{Subtype} & \textbf{Probability} & \textbf{\shortstack{Parameter Range/ \\ Count}} \\
            \midrule
            
            \multirow{3}{*}{\textbf{Defocus Blur}} 
            & \multirow{3}{*}{Gaussian PSF} 
            & Low & 0.30 & $\sigma \in [0.1, 0.5)$ \\
            & & Medium & 0.50 & $\sigma \in [0.5, 1.8)$ \\
            & & High & 0.20 & $\sigma \in [1.8, 2.0)$ \\
            \midrule
            
            \multirow{3}{*}{\textbf{Motion Blur}} 
            & \multirow{3}{*}{Camera Motion Kernel} 
            & $P_1$ (nervous motion) & 0.40 & 540 kernels \\
            & & $P_2$ (oscillatory motion) & 0.36 & 480 kernels \\
            & & $P_3$ (linear motion) & 0.24 & 330 kernels \\
            \midrule
            
            \multirow{3}{*}{\textbf{Geometric Warp}} 
            & \multirow{3}{*}{Surface Deformation} 
            & Simple & 0.33 & 16 meshes \\
            & & Moderate & 0.20 & 5 meshes \\
            & & Complex & 0.47 & 4 meshes \\
            \midrule

            \multirow{4}{*}{\textbf{Non-uniform illumination}} 
            & \multirow{2}{*}{Environmental lighting} & \multirow{2}{*}{HDR Map (Image-based)} & \multirow{2}{*}{0.7} & Rotation: $[0,2\pi)$ rad\\
            & & & & Intensity: 40\%-60\% of metadata \\ \cmidrule(lr){5-5}
            & \multirow{2}{*}{Artificial lighting} & \multirow{2}{*}{Point lap (radial)} & \multirow{2}{*}{0.3} & Distance: 3 to 8 units\\
            & & & & Color temperature: 4000K to 5500K \\
            \bottomrule
            \end{tabular}
        }
    \end{minipage}
\end{figure*}

\begin{figure*}[h]
  \centering
  \includegraphics[width=\textwidth, height=6.2cm]{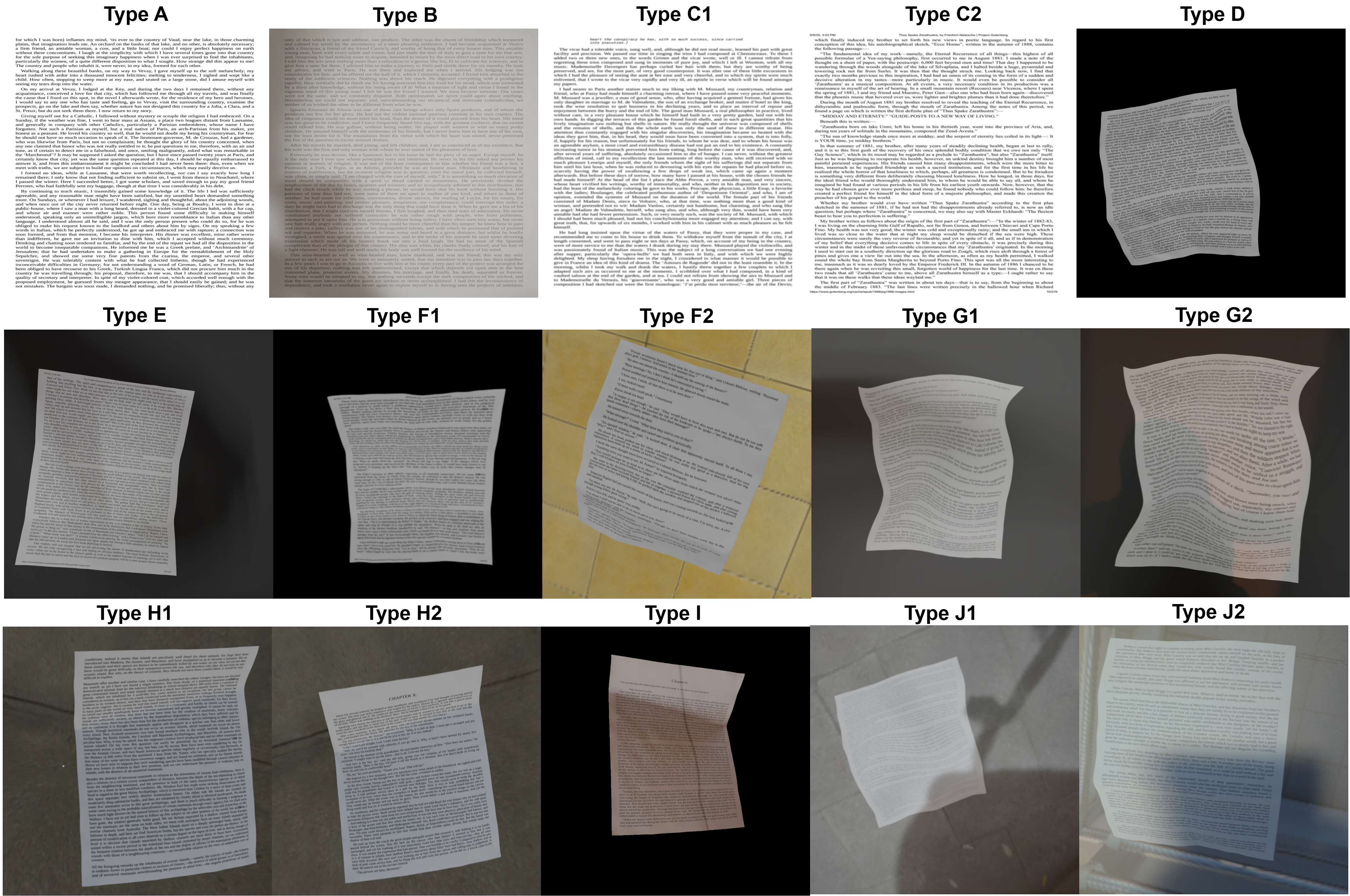}
  \caption{GutenDoc dataset example with corresponding distortion type.(See Table \ref{GutenDoc_dataset}).}
    \label{GutenDoc_data.drawio}
\end{figure*}

\section{GutenDoc Dataset}
Existing document restoration benchmarks typically evaluate models under isolated degradations such as blur, shadows, non-uniform illumination, or geometric warping \cite{inproceedings, feng2022geometric, 8578350, 10268585, das2019dewarpnet, Verhoeven_2023, 9897217, 10.1145/1816123.1816161, 10.5555/1768409.1768429} implicitly assuming that distortions occur independently and can be modeled as separate perturbations applied to a clean image. However, real document images result from a capture process where page geometry, camera motion, and illumination are strongly coupled. Page curvature alters focus and depth-of-field, surface orientation determines shadow formation and lighting falloff, and camera motion introduces directionally correlated motion blur that depends on viewing angle — all of which jointly interact to create compound degradations. Consequently, performance measured under independent corruptions does not reflect real-world robustness.  We therefore formulate document restoration as a compound degradation problem and construct a benchmark designed to emulate the intricate entanglement of simultaneous geometric and photometric distortions.

\subsection{Dataset Construction}
We construct dense text document pages using English novels from the Gutenberg public domain corpus\footnote{\url{https://www.gutenberg.org}}. Unlike prior document restoration datasets that commonly operate at \(256 \times 256\) resolution \cite{zhang2024docresgeneralistmodelunifying, zhao2025unidocdiffunifieddocumentrestoration, zamir2022restormer, das2019dewarpnet}, GutenDoc uses high-resolution \(1344 \times 1344\) images. Preserving high-frequency character strokes prevents ambiguity between text structure and blur, enabling meaningful character-level evaluation, (See Figure ~\ref{1344vs256.drawio}).

We model the document capture pipeline through sequential physical effects: geometry deformation, illumination variation, shadow, and optical blur. Document surfaces are deformed using meshes for realistic page manipulations (warps, perspective distortions, scaling, rotation, and translation), categorized into simple, moderate, and complex severities. Some meshes are adapted from Doc3D \cite{das2019dewarpnet}.

Photometric effects include non-uniform illumination (via a custom Blender\footnote{\url{https://www.blender.org}} setup for providing environmental and artificial lighting), realistic shadows incorporated using patterns adapted from DocAligner \cite{zhang2025docaligner}, and blur. We model two blur types: defocus blur using Gaussian kernels with low (\(\sigma\in[0.1,0.5)\)), medium (\(\sigma\in[0.5,1.8)\)), and high (\(\sigma\in[1.8,2.0)\)) severity \cite{s18041135, s150100880}; and motion blur using camera motion kernels for nervous ($P_1$), oscillatory ($P_2$), and approximately linear ($P_3$) motions with Exposure factor $E_1$ following \cite{Sayed_2021_CVPR}.

The ordering of degradations follows the physical image formation process. Geometric deformation is applied first to define the intrinsic 3D physical state of the document (warp). Next, lighting variations and shadow formations are applied to simulate environmental illumination interacting with that specific 3D geometry. Finally, defocus or motion blur is applied, representing camera sensor limitations during capture. This sequence ensures that all scene elements—including the deformations and their resulting shadows—are blurred uniformly, yielding naturally correlated real-world physical distortions.

\begin{figure*}[t]
    \centering
    \begin{minipage}{0.56\textwidth}
        \centering
        \captionof{table}{GutenDoc: Breakdown of distortion types and dataset distribution.}
        \label{GutenDoc_dataset}
        \renewcommand{\arraystretch}{2}
        \setlength{\tabcolsep}{2pt} 
        \resizebox{\linewidth}{!}{%
        \begin{tabular}{@{}ll cccccc ccc@{}}
        \toprule
        \multirow{3}{*}{\textbf{\shortstack{Distortion\\Count}}} & 
        \multirow{3}{*}{\textbf{Type}} & 
        \multicolumn{6}{c}{\textbf{Distortions present}} & 
        \multirow{3}{*}{\textbf{\shortstack{No. of\\Training\\images}}} & 
        \multirow{3}{*}{\textbf{\shortstack{No. of\\Validation\\images}}} & 
        \multirow{3}{*}{\textbf{\shortstack{No. of\\Testing\\images}}} \\ 
        \cmidrule(lr){3-8}
         & & & \multicolumn{4}{c}{\textbf{Photometric distortions}} & \textbf{Geometric} & & & \\ 
        \cmidrule(lr){4-7} \cmidrule(lr){8-8}
         & & \textbf{None} & \textbf{Shadow} & \textbf{\shortstack{Defocus\\blur}} & \textbf{\shortstack{Motion\\blur}} & \textbf{\shortstack{Non-uniform\\Lighting}} & \textbf{Warps} & & & \\ 
        \midrule
        
        \textbf{0} & \textbf{A} & \textbf{\checkmark} & \textbf{-} & \textbf{-} & \textbf{-} & \textbf{-} & \textbf{-} & 342 & 49 & 19 \\ 
        \midrule
        
        \multirow{4}{*}{\textbf{1}} 
          & \textbf{B}  & \textbf{-} & \textbf{\checkmark} & \textbf{-} & \textbf{-} & \textbf{-} & \textbf{-} & 228 & 32 & 12 \\
          & \textbf{C1} & \textbf{-} & \textbf{-} & \textbf{\checkmark} & \textbf{-} & \textbf{-} & \textbf{-} & 228 & 32 & 12 \\
          & \textbf{C2} & \textbf{-} & \textbf{-} & \textbf{-} & \textbf{\checkmark} & \textbf{-} & \textbf{-} & 228 & 32 & 12 \\
          & \textbf{D}  & \textbf{-} & \textbf{-} & \textbf{-} & \textbf{-} & \textbf{-} & \textbf{\checkmark} & 228 & 32 & 12 \\ 
        \midrule
        
        \multirow{3}{*}{\textbf{2}} 
          & \textbf{E}  & \textbf{-} & \textbf{\checkmark} & \textbf{-} & \textbf{-} & \textbf{-} & \textbf{\checkmark} & 684 & 98 & 38 \\
          & \textbf{F1} & \textbf{-} & \textbf{-} & \textbf{\checkmark} & \textbf{-} & \textbf{-} & \textbf{\checkmark} & 684 & 98 & 38 \\
          & \textbf{F2} & \textbf{-} & \textbf{-} & \textbf{-} & \textbf{\checkmark} & \textbf{-} & \textbf{\checkmark} & 684 & 98 & 38 \\ 
        \midrule
        
        \multirow{5}{*}{\textbf{3}} 
          & \textbf{G1} & \textbf{-} & \textbf{\checkmark} & \textbf{\checkmark} & \textbf{-} & \textbf{-} & \textbf{\checkmark} & 912 & 130 & 51 \\
          & \textbf{G2} & \textbf{-} & \textbf{\checkmark} & \textbf{-} & \textbf{\checkmark} & \textbf{-} & \textbf{\checkmark} & 912 & 130 & 51 \\
          & \textbf{H1} & \textbf{-} & \textbf{-} & \textbf{\checkmark} & \textbf{-} & \textbf{\checkmark} & \textbf{\checkmark} & 912 & 130 & 51 \\
          & \textbf{H2} & \textbf{-} & \textbf{-} & \textbf{-} & \textbf{\checkmark} & \textbf{\checkmark} & \textbf{\checkmark} & 912 & 130 & 51 \\
          & \textbf{I}  & \textbf{-} & \textbf{\checkmark} & \textbf{-} & \textbf{-} & \textbf{\checkmark} & \textbf{\checkmark} & 912 & 130 & 51 \\ 
        \midrule
        
        \multirow{2}{*}{\textbf{4}} 
          & \textbf{J1} & \textbf{-} & \textbf{\checkmark} & \textbf{\checkmark} & \textbf{-} & \textbf{\checkmark} & \textbf{\checkmark} & 5,472 & 788 & 315 \\
          & \textbf{J2} & \textbf{-} & \textbf{\checkmark} & \textbf{-} & \textbf{\checkmark} & \textbf{\checkmark} & \textbf{\checkmark} & 5,472 & 788 & 315 \\ 
        \midrule
        
        \multicolumn{8}{r}{} & \textbf{Total = 18,810} & \textbf{Total = 2,697} & \textbf{Total = 1,066} \\ 
        \bottomrule
        \end{tabular}%
        }
    \end{minipage}
    \hfill 
    \begin{minipage}{0.4\textwidth}
        \centering
        \includegraphics[width=\linewidth]{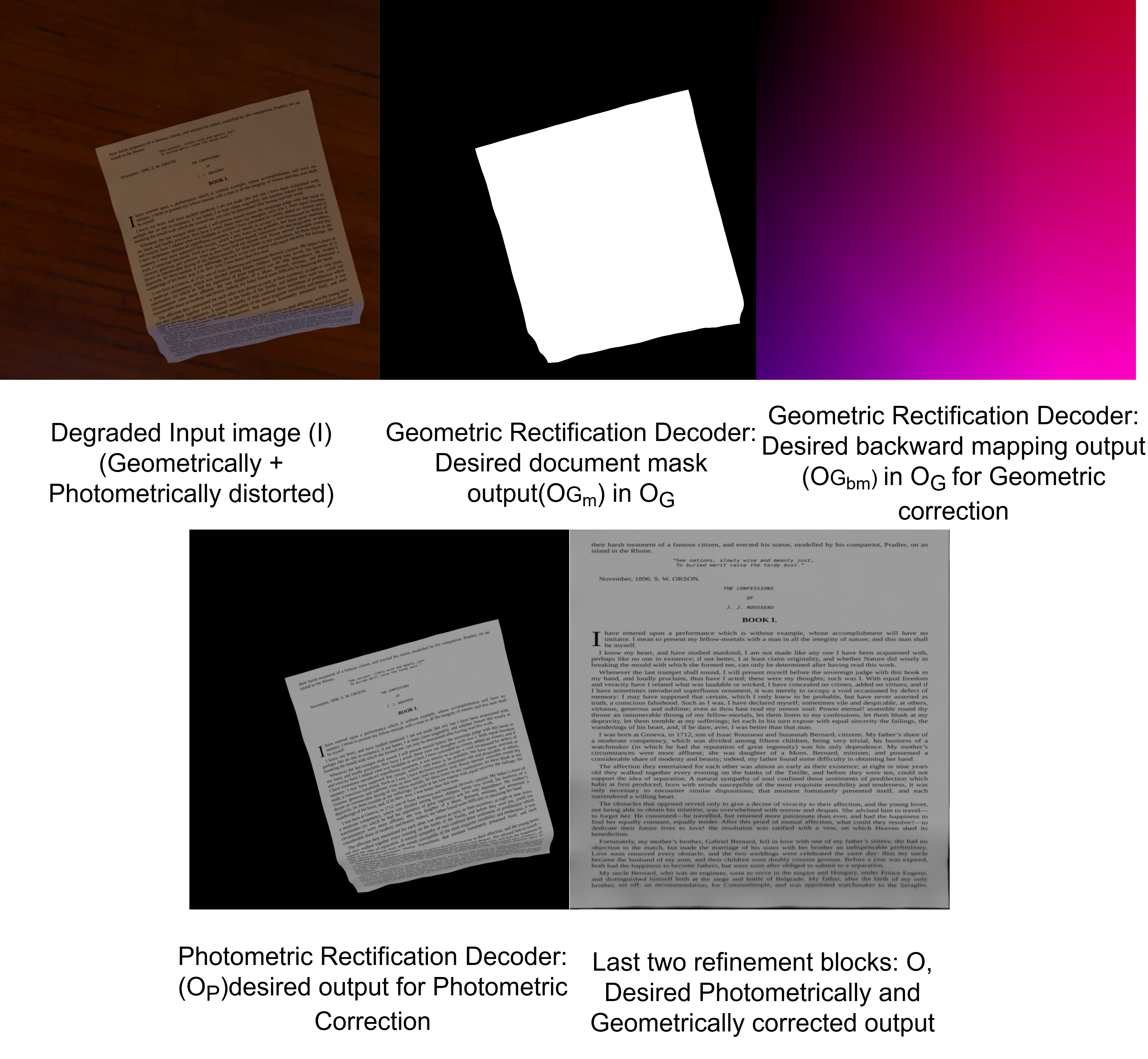}
        \caption{The degraded input image and the desired outputs for ENet-GP at: Geometric Rectification Decoder, Photometric Rectification Decoder, and the last two refinement blocks.}
        \label{ENet-GP_various_stages_input}
    \end{minipage}
\end{figure*}

For each distorted observation we provide ground truth supervision: the distortion-free document image, backward mapping for geometric rectification, and a valid document mask which is a binary image showing the document region within the complete image. Figure \ref{GutenDoc_data.drawio} depicts examples from GutenDoc dataset. The dataset composition and distortion distribution are summarized in Table~\ref{GutenDoc_dataset}, and the parameter statistics in Table~\ref{tab:distortion_params}.


\begin{figure*}[!t]
  \centering
  \includegraphics[width=\textwidth]{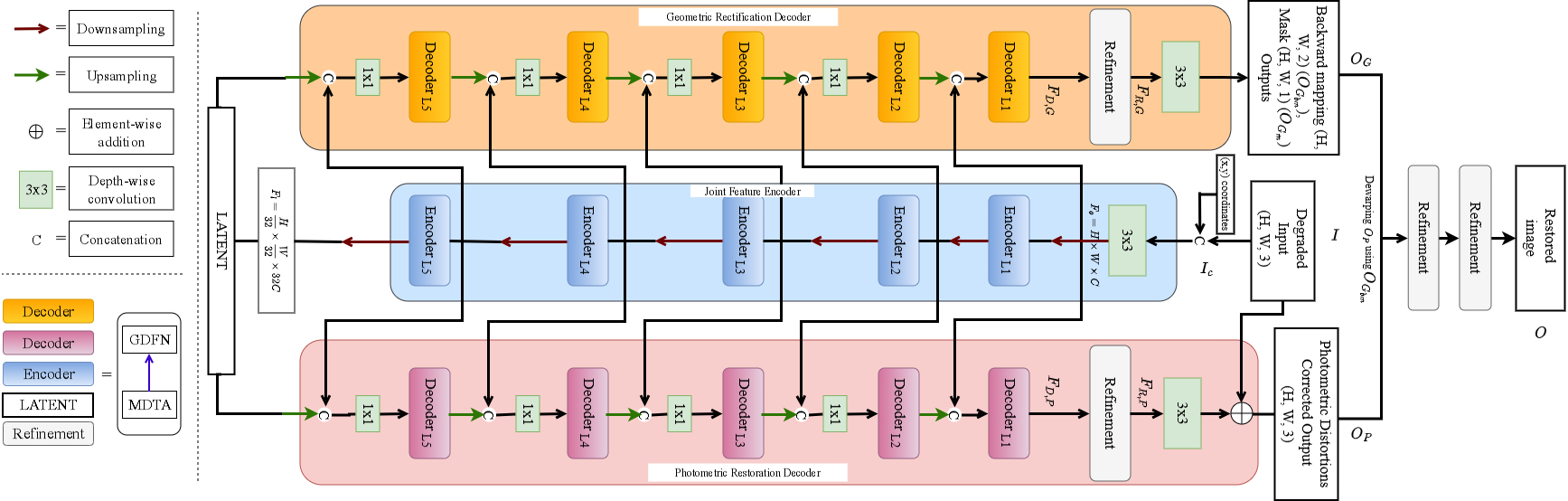}
  \caption{The proposed architecture, ENet-GP. The Joint Feature Encoder (the middle arm) captures features for both photometric and geometric distortions. Geometric Rectification decoder (the top arm) produces backward mapping and document mask region of the input image to solve for only geometric degradations. Photometric Rectification Decoder (the bottom arm) produces only photometric degradation free image inside the document mask region. These two outputs are combined to produce the final geometric and photometric distortion free image.}
  \label{fig:pipeline_arch}
\end{figure*}

\section{Unified Network Architecture}
Our goal is to develop a unified architecture, that can restore any combination of geometric and photometric degradations present in a high-resolution document image in a single step as shown in bottom figure of Figure \ref{fig:overall_contri}. To address the limitations of existing unified solvers, we utilize Restormer \cite{zamir2022restormer}, which has emerged as a standard for image restoration due to its architectural simplicity, computational efficiency, and robustness on high-resolution images. It employs Multi-Dconv Head Transposed Attention (MDTA) to efficiently aggregate local and non-local pixel interactions, and a Gated-Dconv Feed-Forward Network (GDFN) to perform controlled feature transformation by selectively enhancing informative activations. Leveraging these capabilities, our proposed framework is designed as a single-network-single-training setup to solve geometric and photometric degradations simultaneously, avoiding sequential configurations (as shown in Figure \ref{fig:overall_contri}). By harnessing these mechanisms to jointly correct spatial and intensity-based distortions, our model reliably transforms casually captured images into high-fidelity outputs. The architecture is shown in Figure \ref{fig:pipeline_arch}.

The input to the network is the degraded input document image \( I \in \mathbb{R}^{H \times W \times 3} \) (\(H\) and \(W\) denotes the spatial dimension, \(H = 1344, W = 1344\)), which suffers from both geometric and photometric degradations. The image \(I \) is concatenated with the $x$ and $y$ coordinates of the pixel at \((i, j)\), i.e. \(P_{cx}(i, j) = i\), and \(P_{cy}(i, j) = j\), to give \( I_c \in \mathbb{R}^{H \times W \times 5} \). This facilitates backward mapping learning, by giving a better perception of positional information, inspired by \cite{das2019dewarpnet, liu2018intriguingfailingconvolutionalneural}.

\textbf{Joint Feature Encoder} transforms \(I_c\) to latent features \( F_l \in \mathbb{R}^{\frac{H}{32}\times \frac{W}{32} \times 32C} \), where \( C \) is the number of channels (\(C = 12\)). First, a 3x3 convolution kernel extracts initial low-level features \( F_0 \in \mathbb{R}^{H \times W \times C} \) from \(I_c\). \(F_0 \) then passes through the 6-level of joint feature encoder which captures features for both photometric and geometric distortions. The features are iteratively downsampled by a factor 2, using pixel-unshuffle \cite{shi2016realtimesingleimagevideo}, leading to the latent features \( F_l \in \mathbb{R}^{\frac{H}{32}\times \frac{W}{32} \times 32C} \). 

The number of Transformer blocks are gradually increased from top (L1 level of encoder and decoder) to bottom (Latent block) levels to maintain efficiency. Symmetric 6-level geometric rectification and photometric rectification decoders independently process \(F_l\) into \( F_{D,G} \in \mathbb{R}^{H \times W \times 2C}\) and \( F_{D,P} \in \mathbb{R}^{H \times W \times 2C}\), respectively. At each decoder level and latent block, features are upsampled by a factor of 2, using pixel shuffle \cite{shi2016realtimesingleimagevideo}, and concatenated with corresponding encoder features via skip connections \cite{ronneberger2015unetconvolutionalnetworksbiomedical} to mitigate vanishing gradients. A 1x1 convolution then halves the channel dimension following each concatenation. (Refer to Figure \ref{ENet-GP_various_stages_input} for the inputs and outputs used by the architecture for supervision).
 
\textbf{Geometric Rectification Decoder} outputs \( O_G \in \mathbb{R}^{H \times W \times 3}\) after passing \( F_{D,G} \in \mathbb{R}^{H \times W \times 2C}\) through a refinement block to produce \( F_{R,G} \in \mathbb{R}^{H \times W \times 2C}\) to enrich its features which is then passed through a 3x3 convolution to reduce the channels from \(C = 12\) to \(C =3\). \(O_G\) consists of a 2D backward mapping \(O_{G_{bm}} \in \mathbb{R}^{H \times W \times 2}\) to dewarp the image and the 1D binary document mask \(O_{G_m} \in \mathbb{R}^{H \times W \times 1}\) defining the document region within the background. Learning this mask helps the network capture the global document layout, which improves photometric correction. $L_1$ loss is used between \(O_{G_{bm}}\) and its corresponding ground truth \(O_{G_{bm}}^{GT} \in \mathbb{R}^{H \times W \times 2}\), and \(O_{G_m}\) and its corresponding ground truth \(O_{G_{m}}^{GT} \in \mathbb{R}^{H \times W \times 1}\) (see Figure \ref{ENet-GP_various_stages_input}). The sum of these losses \(L_{geo}\) (Equation \ref{equation_1}) is used to update the geometric decoder arm and the joint feature encoder. 

\begin{gather}
\label{equation_1}
\mathcal{L}_{geo} = \mathcal{L}_{bm} + \mathcal{L}_{m}; \quad \mathcal{L}_{bm} = \| O_{G_{bm}} - O_{G_{bm}}^{GT} \|_1, \quad \mathcal{L}_{m} = \| O_{G_{m}} - O_{G_{m}}^{GT} \|_1 \\
\label{equation_2}
\mathcal{L}_{pho} = \quad \| O_{P} - O_P^{GT} \|_1 \\
\label{equation_3}
\mathcal{L}_{crsip} = \quad \| O - O^{GT} \|_1 \\
\label{equation_4}
\mathcal{L}_{total} = \mathcal{L}_{geo} + \mathcal{L}_{pho} + \mathcal{L}_{crisp}
\end{gather}

\textbf{Photometric Rectification Decoder} outputs \( O_P \in \mathbb{R}^{H \times W \times 3}\) after passing \( F_{D,P} \in \mathbb{R}^{H \times W \times 2C}\) through a refinement block to produce \( F_{R,P} \in \mathbb{R}^{H \times W \times 2C}\) to enrich its features, which is then passed through a 3x3 convolution to reduce the channels from \(C = 12\) to \(C = 3\), and then adding it element-wise with \(I \in \mathbb{R}^{H \times W \times 3}\). \(O_P\) represents only the photometrically corrected output in the document mask region that is free from blurs, shadows, and non-uniform illumination. It still suffers from geometric distortions though i.e. warps. \(L_{pho}\) (Equation \ref{equation_2})is the $L_1$ loss between \(O_P\) and its corresponding ground truth \(O_P^{GT} \in \mathbb{R}^{H \times W \times 3}\) for training (refer Figure \ref{ENet-GP_various_stages_input}), and is used to update the photometric decoder and the joint feature encoder. 

To produce the restored image output $O \in \mathbb{R}^{H \times W \times 3}$, the backward mapping from $O_G$ is used to dewarp $O_P$. This intermediate result passes through two Refinement blocks to generate a high-resolution image, completely free of both geometric and photometric distortions. \(L_{crisp}\) (Equation \ref{equation_3})is the $L_1$ loss between $O$ and its corresponding ground truth \(O^{GT} \in \mathbb{R}^{H \times W \times 3}\) (refer Figure \ref{ENet-GP_various_stages_input}) for updating these refinement blocks.

The total loss (\(L_{total}\)) (Equation \ref{equation_4}) is the sum of \(L_{geo}\) (Equation \ref{equation_1}), \(L_{pho}\) (Equation \ref{equation_2}), and \(L_{crisp}\) (Equation \ref{equation_3}), as we consider each loss to be equally likely to influence the restored image output, used during single pass backward propagation.

The document mask from the geometric decoder is not utilized during inference. Our approach generalizes better in complex real-world scenarios and avoids the high latency of iterative generative diffusion models \cite{zhang2025dvdunleashinggenerativeparadigm, yang2023docdiffdocumentenhancementresidual}, while delivering competitive performance.


\begin{table}
\centering
\caption{Comparison of ENet-GP and DocRes \cite{zhang2024docresgeneralistmodelunifying} on GutenDoc testing dataset.  For each metric, the best result achieved across a group and its associated prime rows (row $A$ and rows of $A^\prime$, row $B$ and rows of $B^\prime$) is highlighted in \textbf{bold}.“$\downarrow$” means lower is better. “$\uparrow$” means higher is better.}
\label{tab:dataset_claim}
\resizebox{\textwidth}{!}{%
\begin{tabular}{@{}c l l l c cccc cc cc cc@{}}
\toprule
\multirow{4}{*}{\textbf{Combination}} & \multirow{4}{*}{\textbf{Model}} & \multirow{4}{*}{\textbf{\shortstack{Trained\\on}}} & \multirow{4}{*}{\textbf{Group}} & \multirow{4}{*}{\textbf{\shortstack{Tested\\on}}}  & \multicolumn{4}{c}{\textbf{Inference Passes}} & \multicolumn{2}{c}{\textbf{Image Metrics}} & \multicolumn{2}{c}{\textbf{Text Metrics}} & \multirow{4}{*}{\textbf{\shortstack{Average\\Wall-clock\\Time (s)$\downarrow$}}} & \multirow{4}{*}{\textbf{GFLOPs$\downarrow$}} \\ 
\cmidrule(lr){6-9} \cmidrule(lr){10-11} \cmidrule(lr){12-13}
 & & & & & \multirow{3}{*}{\textbf{1st}} & \multirow{3}{*}{\textbf{2nd}} & \multirow{3}{*}{\textbf{3rd}} & \multirow{3}{*}{\textbf{4th}} & \multirow{2}{*}{\textbf{PSNR$\uparrow$/}} & \multirow{2}{*}{\textbf{SSIM$\uparrow$/}} & \multirow{3}{*}{\textbf{CER$\downarrow$}} & \multirow{3}{*}{\textbf{ED$\downarrow$}} & &  \\ 
 & & & & & & & & & & & & & \\ 
\midrule

\multirow{6}{*}{1} & ENet-GP & \multirow{6}{*}{\shortstack{GutenDoc\\(Train and\\Val split)\\(see Table \ref{GutenDoc_dataset})}} & $A\rightarrow$ & GutenDoc Test split (Table \ref{GutenDoc_dataset})  & Perspective All & -- & -- & -- & 22.1313 & 0.6551 & \textbf{0.5118} & \textbf{1493.8311} & \textbf{0.439} & 280.2937 \\ \cmidrule(lr){2-2}  \cmidrule(lr){4-15}
 & \multirow{5}{*}{DocRes} & & \multirow{5}{*}{$A^\prime\rightarrow$}\ldelim\{{5}{1pt} & \multirow{5}{*}{GutenDoc Test split (Table \ref{GutenDoc_dataset})} & Dew & -- & -- & -- & 16.4945 & 0.6309 & 0.8622 & 2581.0497 & 1.5326 & 183.6135 \\
 & & & & & Dew & Des & -- & -- & 22.5562 & 0.7309 & 0.9767 & 2928.8002 & 1.9852 & 367.227 \\
 & & & & & Dew & App & -- & -- & 22.5499 & 0.7307 & 0.9738 & 2917.6717 & 1.6372 & 367.227 \\
 & & & & & Dew & Deb & -- & -- & \textbf{22.6066} & 0.7307 & 0.9768 & 2928.8490 & 1.9092 & 367.227 \\
 & & & & & Dew & Deb & Des & App & 22.2360 & \textbf{0.7316} & 0.9967 & 2931.1341 & 3.8834 & 734.454 \\
\bottomrule
\end{tabular}%
}
\end{table}

\begin{table}
\centering
\caption{Comparison of ENet-GP and DocRes \cite{zhang2024docresgeneralistmodelunifying} on individual task-specific testing dataset of IDD.  For each metric, the best result achieved across a group and its associated prime row (e.g. $C$ and $C^\prime$ row, $D$ and $D^\prime$ row,...)  is highlighted in \textbf{bold}.“$\downarrow$” means lower is better. “$\uparrow$” means higher is better.}
\label{tab:ENet-GP_arch_claim}
\resizebox{\textwidth}{!}{%
\begin{tabular}{@{}c l l l c c cc cc cc@{}}
\toprule
\multirow{4}{*}{\textbf{Combination}} & \multirow{4}{*}{\textbf{Model}} & \multirow{4}{*}{\textbf{\shortstack{Trained\\on}}} & \multirow{4}{*}{\textbf{Group}} & \multirow{4}{*}{\textbf{\shortstack{Tested\\on}}}  & \multicolumn{1}{c}{\textbf{Inference Passes}} & \multicolumn{2}{c}{\textbf{Image Metrics}} & \multicolumn{2}{c}{\textbf{Text Metrics}} & \multirow{4}{*}{\textbf{\shortstack{Average\\Wall-clock\\Time (s)$\downarrow$}}} & \multirow{4}{*}{\textbf{GFLOPs$\downarrow$}} \\ 
\cmidrule(lr){6-6} \cmidrule(lr){7-8} \cmidrule(lr){9-10}
 & & & & & \multirow{3}{*}{\textbf{1st}} & \multirow{2}{*}{\textbf{\shortstack{\textbf{PSNR$\uparrow$/}\\\textbf{(LD)$\downarrow$}}}} & \multirow{2}{*}{\textbf{\shortstack{\textbf{SSIM$\uparrow$/}\\\textbf{(MSSIM)$\uparrow$}}}} & \multirow{3}{*}{\textbf{CER$\downarrow$}} & \multirow{3}{*}{\textbf{ED$\downarrow$}} & & \\ 
 & & & & & & & & & & & \\ 
\midrule

\multirow{14}{*}{2} & \multirow{7}{*}{ENet-GP} & \multirow{14}{*}{\shortstack{GutenDoc\\(Train and\\Val split)\\(see Table \ref{GutenDoc_dataset})}} 
 & $J \rightarrow$ & DIR300 \cite{feng2022geometric} & Perspective All & (15.9966) & (0.4980) & \textbf{0.4951} & 1250.4833 & \textbf{1.4466} & 280.2937 \\ 
 & & & $K \rightarrow$ & Kligler \cite{8578350} & Perspective All & 6.6137 & 0.3719 & \textbf{0.5431} & \textbf{115.7933} & \textbf{0.4660} & 280.2937 \\
 & & & $L \rightarrow$ & Jung \cite{jung2019waterfillingefficientalgorithmdigitized} & Perspective All & \textbf{15.2150} & \textbf{0.7626} & 0.9537 & \textbf{545.5747} & \textbf{0.4660} & 280.2937 \\
 & & & $M \rightarrow$ & OSR \cite{s20236929} & Perspective All & 12.5555 & \textbf{0.8549} & \textbf{0.2662} & \textbf{299.9451} & \textbf{0.4660} & 280.2937 \\ 
 & & & $N \rightarrow$ & TDD \cite{inproceedings} & Perspective All & 7.9188 & 0.2823 & 0.8843 & 136.3169 & 0.4167 & 280.2937 \\ 
 & & & $O \rightarrow$ & RealDAE \cite{10268585} & Perspective All & \textbf{16.3804} & \textbf{0.7621} & \textbf{0.6967} & \textbf{249.5067} & \textbf{0.9186} & 280.2937 \\
 & & & $P \rightarrow$ & DocUNet* \cite{Ma_2018_CVPR} & Perspective All & \textbf{15.6968} & \textbf{0.7161} & \textbf{0.2973} & \textbf{569.1} & \textbf{0.9186} & 280.2937 \\ \cmidrule(lr){2-2} 
  \cmidrule(lr){4-12}
 & \multirow{7}{*}{DocRes} & & $J^\prime \rightarrow$ & DIR300 \cite{feng2022geometric} & Dew  & \textbf{(11.3754)} & \textbf{(0.5351)} & 0.5509 & \textbf{1152.8267} & 3.3939 & 183.6135 \\ 
 & & & $K^\prime \rightarrow$ & Kligler \cite{8578350} & Des  & \textbf{17.2751} & \textbf{0.7774} & 0.5699 & 140.1933 & 1.2208 & 183.6135 \\
 & & & $L^\prime \rightarrow$ & Jung \cite{jung2019waterfillingefficientalgorithmdigitized} & Des  & 11.6288 & 0.5353 & \textbf{0.8431} & 915.954 & 1.2208 & 183.6135 \\
 & & & $M^\prime \rightarrow$ & OSR \cite{s20236929} & Des  & \textbf{14.0855} & 0.7108 & 0.5115 & 548.9030 & 1.2208 & 183.6135 \\ 
 & & & $N^\prime \rightarrow$ & TDD \cite{inproceedings} & Deb  & \textbf{11.0158} & \textbf{0.5332} & \textbf{0.7419} & \textbf{110.56} & 0.0472 & 183.6135 \\ 
 & & & $O^\prime \rightarrow$ & RealDAE \cite{10268585} & App  & 8.7246 & 0.6838 & 0.9320 & 524.82 & 1.9168 & 183.6135 \\
 & & & $P^\prime \rightarrow$ & DocUNet* \cite{Ma_2018_CVPR} & App  & 9.8348 & 0.6109 & 0.5423 & 1452.2769 & 1.9168 & 183.6135 \\
\bottomrule
\end{tabular}%
}
\end{table}

\section{Experiments}
We compare with unified models whose training and evaluation code is publicly available. Consequently, we compare against DocRes\cite{zhang2024docresgeneralistmodelunifying}, but exclude \cite{zhao2025unidocdiffunifieddocumentrestoration, 10447446} due to lack of accessible source code.  

Datasets and Metrics: We evaluate our model using two data regimes: the GutenDoc dataset, which contains high-resolution dense text document images with complex and entangled distortions making it a Multiple-distortion dataset (MDD), and various Isolated-distortion datasets (IDD) (from Table \ref{tab:related_works_datasets}). Image restoration quality is quantified using Peak Signal-to-Noise Ratio (PSNR) and Structural Similarity Index (SSIM). For the only dewarping task, we report Local Distortion (LD) \cite{you2016multiviewrectificationfoldeddocuments} and Multi-Scale SSIM (MSSIM) \cite{1292216}. Downstream OCR performance is assessed via Character Error Rate (CER) and Edit Distance (ED), using Tesseract \cite{4376991}. To benchmark computational efficiency, we report average wall-clock inference time (loading + inference + saving) per testing image, in seconds, for corresponding testing datasets on NVIDIA A100 80GB GPU, and GFLOPs.

IDD: We refer to the following collection of individual task-specific document restoration datasets \cite{zhang2024docresgeneralistmodelunifying, zhao2025unidocdiffunifieddocumentrestoration, Verhoeven_2023, feng2022doctrdocumentimagetransformer, inproceedings, yang2023docdiffdocumentenhancementresidual} as IDD. 
\begin{itemize}
    \item Dewarping (Dew): Training uses Doc3D \cite{das2019dewarpnet} (100K synthetic samples); testing is performed on the real-world DIR300 \cite{feng2022geometric}.
    \item Deshadowing (Des): Training uses 14,200 synthetic images from FSDSRD \cite{9897217} and 4,371 real images from RDD \cite{Zhang_2023_CVPR}. Testing is conducted on Jung \cite{jung2019waterfillingefficientalgorithmdigitized}, Kligler \cite{8578350}, and OSR \cite{s20236929}.
    \item Deblurring (Deb): Training utilizes 40,000 samples from the Text Deblur Dataset (TDD) \cite{inproceedings}, with 1,600 samples reserved for testing.
    \item Appearance Enhancement (App): Training combines 90,000 images from Doc3DShade \cite{das2020intrinsicdecompositiondocumentimages} and 450 images from RealDAE \cite{10268585}. Evaluation is performed on RealDAE, and DocUNet \cite{Ma_2018_CVPR} after alignment using DocAligner \cite{zhang2025docaligner}.
\end{itemize}

We report performance by training using the GutenDoc training and validation set (as in Table \ref{GutenDoc_dataset}). For evaluation on testing set of GutenDoc, which is MDD, we employ DocRes \cite{zhang2024docresgeneralistmodelunifying} with multiple combinations of passes in a sequential cascade manner (see bottom of Figure\ref{fig:overall_contri} and Table \ref{tab:dataset_claim}). The sequence ordering is arbitrarily chosen, keeping dewarping first, so as to establish a stable document region coordinate system. We refer to the corresponding sequence of DocRes in $`$Inference passes$'$ of Table \ref{tab:dataset_claim} as DocRes(\(Dew_1\)), DocRes(\(Dew_1\), \(Des_2\)), DocRes(\(Dew_1\), \(App_2\)), DocRes(\(Dew_1\), \(Deb_2\)), and DocRes(\(Dew_1\), \(Deb_2\), \(Des_3\), \(App_4\)) correspondingly. We also use the individual task-specific testing dataset in IDD for evaluation (as in Table \ref{tab:ENet-GP_arch_claim}). For Table \ref{tab:ENet-GP_arch_claim} only one inference pass is used for DocRes\cite{zhang2024docresgeneralistmodelunifying} as the task for each IDD testing image is known beforehand, so corresponding required single pass can be done. In Table \ref{tab:dataset_claim} and \ref{tab:ENet-GP_arch_claim}, $'$Perspecive All$'$ indicates that ENet-GP infers on an image considering it suffers from all the distortion that were used for training, regardless of whether the image has single or multiple distortions present. 

As geometric deformations significantly reduce quality of OCR text metrics \cite{6909892, istomin2025efficientdocumentimagedewarping}, we present text metrics comparing ENet-GP with recent works, as detailed in Tables \ref{tab:dir300_text_results} and \ref{tab:docunet_dewarping_text_results}. For this case, training and testing dataset includes only the $`$dewarping$'$ subset of IDD, as done in \cite{feng2022geometric}.

For evaluation using aggregate of training dataset of IDD and other combinations of $`$Inference passes$'$ of DocRes\cite{zhang2024docresgeneralistmodelunifying}, refer to supplementary.

\subsection{Implementation details}
For every metric of ENet-GP in Tables \ref{tab:dataset_claim}, \ref{tab:ENet-GP_arch_claim}, \ref{tab:dir300_text_results}, \ref{tab:docunet_dewarping_text_results}, the network was trained for 50 epochs using 8 NVIDIA RTX 6000 Ada Generation GPUs (total batch size 8). We utilized a cosine annealing scheduler decaying from $2 \times 10^{-4}$ to $1 \times 10^{-6}$ with restarts every 10,000 iterations. 

Training on GutenDoc (MDD): Required inputs and outputs are selected for each image as per the requirement of DocRes \cite{zhang2024docresgeneralistmodelunifying} for doing unified training, and our network (see Figure \ref{ENet-GP_various_stages_input}).

Training on $`$dewarping$'$ subset of IDD (for Tables \ref{tab:dir300_text_results} and \ref{tab:docunet_dewarping_text_results}): To facilitate unified training on $`$dewarping$'$ subset of IDD having only geometric distortion, we synthesized the missing input for each image, as required by ENet-GP's  architecture (Figure \ref{ENet-GP_various_stages_input}). We provide a binary document mask ground truth \(O_{G_{m}}^{GT}\) and set the photometric ground truth \(O_{P}^{GT}\) equal to the degraded input $I$. The restored image output $O$ is supervised by applying the ground-truth backward mapping to $I$.

\begin{table}
    \centering
    \small
    
    \begin{minipage}{0.45\textwidth}
        \centering
        \caption{Quantitative comparisons of existing learning-based methods trained on $`$dewarping$'$ subset of IDD in terms of OCR accuracy on the DIR300 \cite{feng2022geometricrepresentationlearningdocument} test set. “$\downarrow$” means lower if better.}
        \label{tab:dir300_text_results}
        \begin{tabular}{c|cc}
            \hline
            Methods & ED $\downarrow$ & CER $\downarrow$ \\ \hline
            Distorted & 1500.56 & 0.5234 \\
            DocProj \cite{10.1145/3355089.3356563} & 958.89 & 0.3540 \\
            DewarpNet \cite{das2019dewarpnet} & 1059.57 & 0.3557 \\
            DocTr \cite{feng2022doctrdocumentimagetransformer} & 699.63 & 0.2237 \\
            DDCP \cite{xie2022documentdewarpingcontrolpoints} & 2084.97 & 0.5410 \\
            DocGeo \cite{feng2022geometric} & 664.96 & 0.2189 \\ \hline
            ENet-GP(Ours) & 1017.11 & 0.3096\\ \hline
        \end{tabular}
        \caption{Ablation model comparison.}
        \label{tab:ENet-GP_ablation}
        \resizebox{\textwidth}{!}{%
        \begin{tabular}{@{}l|c|c|c@{}}
            \toprule
            \textbf{Metric} & \textbf{Ablation0} & \textbf{Ablation1} & \textbf{ENet-GP (Ours)} \\ \midrule
            PSNR $\uparrow$ & 20.9916 & 21.0185 & 22.1313 \\
            SSIM $\uparrow$ & 0.6327 & 0.6343 & 0.6551 \\
            CER $\downarrow$ & 0.5270 & 0.5263 & 0.5118 \\
            ED $\downarrow$ & 1548.3809 & 1524.1812 & 1493.8311 \\ \bottomrule
        \end{tabular}%
        }
    \end{minipage}
    \hfill
    \begin{minipage}{0.52\textwidth}
        \centering
        \caption{Quantitative comparisons of the existing learning-based methods trained on $`$dewarping$'$ subset of IDD in terms of OCR accuracy on the DocUNet Benchmark dataset \cite{Ma_2018_CVPR}. “$\downarrow$” means lower if better.}
        \label{tab:docunet_dewarping_text_results}
        \begin{tabular}{c|cc}
            \hline
            Methods & ED $\downarrow$ & CER $\downarrow$ \\ \hline
            Distorted & 2111.56/1552.22 & 0.5352/0.5089 \\
            DocUNet \cite{Ma_2018_CVPR} & 1933.66/1259.83 & 0.4632/0.3966 \\
            DocProj \cite{10.1145/3355089.3356563} & 1712.48/1165.93 & 0.4267/0.3818 \\
            FCN-based \cite{xie2021dewarpingdocumentimagedisplacement} & 1792.60/1031.40 & 0.4213/0.3156 \\
            DewarpNet \cite{das2019dewarpnet} & 885.90/525.45 & 0.2373/0.2102 \\
            PWUNet \cite{9710331} & 1069.28/743.32 & 0.2677/0.2623 \\
            DocTr \cite{feng2022doctrdocumentimagetransformer} & 724.84/464.83 & 0.1832/0.1746 \\
            DDCP \cite{xie2022documentdewarpingcontrolpoints} & 1442.84/745.35 & 0.3633/0.2626 \\
            FDRNet \cite{9880145} & 829.78/514.90 & 0.2068/0.1846 \\
            RDGR \cite{jiang2022revisiting} & 729.52/420.25 & 0.1717/0.1559 \\ 
            DocGeo \cite{feng2022geometricrepresentationlearningdocument} & 713.94/379.00 & 0.1821/0.1509 \\ \hline
            ENet-GP(Ours) & 1022.76/850.27 &  0.2747/0.2957\\ \hline
        \end{tabular}
    \end{minipage}
\end{table}

\begin{figure*}[!t]
  \centering
  \includegraphics[width=\textwidth, height=12cm]{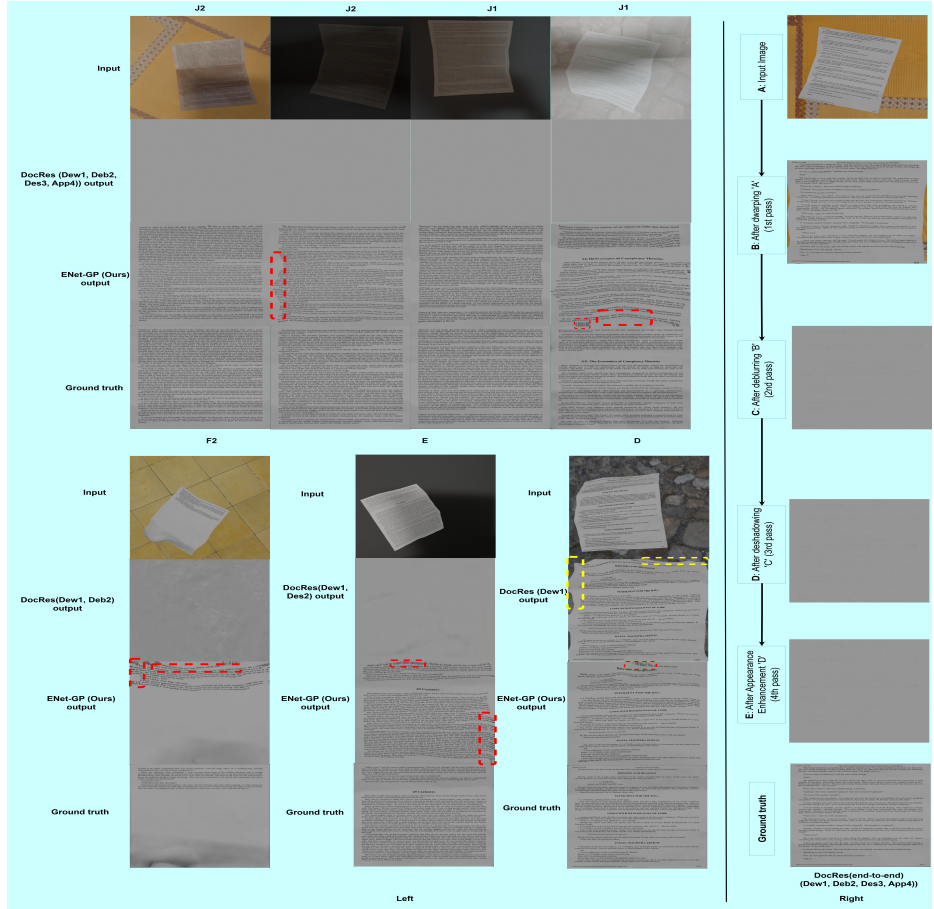}
  \caption{Left: Visual comparison of input image from GutenDoc testing split (Table \ref{GutenDoc_dataset}); restored output from corresponding inference sequence of DocRes \cite{zhang2024docresgeneralistmodelunifying} selected from Table \ref{tab:dataset_claim}; restored output from ENet-GP; and ground truth. Image is washed out for Type J1, J2, D, and E for the corresponding sequential pass of DocRes. ENet-GP output is closely aligned with Ground Truth. Red boxes highlight the regions of stretches, squeeze, and curves in the output. Yellow boxes highlights the background region present in the output. Right: Sequential transformation of image as in 4-inference pass sequences of DocRes(\(Dew_1\), \(Deb_2\), \(Des_3\), \(App_4\)) (from Table \ref{tab:dataset_claim}) which starts with dewarping, then followed by deblurring, deshadowing, and finally appearance enhancement. It depicts the error accumulation as mentioned in \cite{zhang2024docresgeneralistmodelunifying} paper.}
  \label{Qualitative_comparison_+_DocRes_sequential_transformation_final _6}
\end{figure*}

\subsection{Results}
Tables ~\ref{tab:dataset_claim} and \ref{tab:ENet-GP_arch_claim} present evaluation of ENet-GP with DocRes \cite{zhang2024docresgeneralistmodelunifying} on GutenDoc testing split (see Table \ref{GutenDoc_dataset}) and individual task-specific testing dataset in IDD, respectively, when trained using GutenDoc train and val split (from Table\ref{GutenDoc_dataset}).

\textbf{Quantitative comparisons:} In Table \ref{tab:dataset_claim} Combination 1, ENet-GP significantly surpasses the sequences: DocRes(\(Dew_1\)), DocRes(\(Dew_1\), \(Des_2\)), DocRes(\(Dew_1\), \(App_2\)), DocRes(\(Dew_1\), \(Deb_2\)), and DocRes(\(Dew_1\), \(Deb_2\), \(Des_3\), \(App_4\)), consisting of  GutenDoc training and testing dataset, by \textrm{40.64\%} in CER (wrt DocRes(\(Dew_1\))) and by \textrm{42.12\%} in ED (wrt DocRes(\(Dew_1\))). This is achieved while being \textrm{3.5} times faster than DocRes(\(Dew_1\)), which is not even the best in any of PSNR, SSIM, CER, and ED metrics. ENet-GP accomplishes this as a single-pass network while experiencing marginal deviations of less than \textrm{2.2}\% in PSNR (wrt DocRes(\(Dew_1\), \(Deb_2\))) and \textrm{10.5}\% in SSIM (wrt DocRes(\(Dew_1\), \(Deb_2\), \(Des_3\), \(App_4\))). In Table \ref{tab:ENet-GP_arch_claim}, for Combination 2, ENet-GP yields improved CER for the dewarping-task (Group J), deshadowing-task (Group K, M), and Appearance enhancement task (Group O, P). ENet-GP achieves a lower ED for deshadowing-task (Group K, L, M) and Appearance enhancement task (Group O, P). These gains in OCR-based metrics are achieved alongside a substantial reduction in average wall-clock time compared to the DocRes model. This substantiates the claim that ENet-GP is a more versatile, faster and robust unified model than DocRes. The primary limitation, however, is that ENet-GP cannot be manually adjusted to target specific distortions, if known beforehand, during inference. It relies entirely on the parameter optimization achieved during the training phase. Training on a diverse set of entangled distortions drives the model toward doing restoration from a multiple-distortion present perspective, as mostly found in real-word casually captured document photos. ENet-GP performs competitively in Tables \ref{tab:dir300_text_results} and \ref{tab:docunet_dewarping_text_results}. 
 
\textbf{Qualitative comparisons:} Figure \ref{Qualitative_comparison_+_DocRes_sequential_transformation_final _6} (left) shows visual comparisons of input image from GutenDoc testing split (refer Table \ref{GutenDoc_dataset}) suffering with either multiple (Type J1, J2, F2, E) or single (Type D) distortion, the corresponding inference sequence of DocRes \cite{zhang2024docresgeneralistmodelunifying} selected from Table \ref{tab:dataset_claim} for getting restored image, output from ENet-GP, and the ground truth. As Type J1 and J2 has images suffering from 4 distortions: warps, blurs, shadows and non-uniform lighting, 4-inference pass sequence DocRes(\(Dew_1\), \(Deb_2\), \(Des_3\), \(App_4\)) is selected. The restored image for this is completely washed out leading to blank or dull images, which is due to error accumulation (mentioned in \cite{zhang2024docresgeneralistmodelunifying}),  as visible in Figure \ref{Qualitative_comparison_+_DocRes_sequential_transformation_final _6} (Right). As Type F2 and E has images suffering from 2 distortions: warp and blur, and warp and shadow, respectively, the corresponding 2-inference pass sequence, DocRes(\(Dew_1\), \(Deb_2\)) and DocRes(\(Dew_1\), \((Des_2\)) have been selected, respectively, (see Table  \ref{tab:dataset_claim}). The output for these sequences have artifacts and are again washed out. As Type D has images with only one distortion, warp, the corresponding 1-inference pass sequence DocRes(\(Dew_1\)), from Table \ref{tab:dataset_claim} is selected. This restored image has residual background present towards the border (as visible from the yellow highlights).

For the same inputs, Type J1, J2, F2, E and D, ENet-GP outputs image in a single-pass which is closely aligned with ground truth, making ENet-GP a more robust and versatile network. ENet-GP achieves near-perfect deshadowing and deblurring for types J1 and J2. The text towards the border appears squeezed or stretched or curved, as highlighted by red in Type J2, J1, F2, E and D, pointing out that ENet-GP struggles to resolve images with complex warps. 

This also highlights a critical gap in purely quantitative evaluations. For any document restoration model, the evaluation must rely on a combination of image (PSNR, SSIM) and text (CER, ED) metrics to reliably gauge fidelity of restored images and performance for downstream OCR segmentation tasks. Image metrics alone are insufficient without accompanying qualitative evaluation. 

For more Qualitative results, refer to supplementary. 

\subsection{Ablation Studies}

We evaluate the impact of the final two refinement blocks (see Figure \ref{fig:pipeline_arch}) on the restored image output $O$. ENet-GP defaults to two refinement blocks and an identical number of blocks in the encoder and each of the two decoders, based on the premise that restoring photometric and geometric distortions are equally important tasks. We compare this with the following:

Ablation 0 (No Refinement): The restored image output $O \in \mathbb{R}^{H \times W \times 3}$ is generated by dewarping $O_P$ using the backward mapping from $O_G$, without passing through any refinement blocks

Ablation 1 (Single Block): The dewarped output is passed through only one refinement block to generate the restored image output $O \in \mathbb{R}^{H \times W \times 3}$.

As shown in Table \ref{tab:ENet-GP_ablation}, while increasing refinement blocks improves quantitative metrics, it also raises training compute. We retain two blocks as the default to optimize the balance between restoration quality and computational efficiency.

\section{Conclusion}

We introduce GutenDoc dataset, a high-resolution dense text semi-realistic document restoration dataset containing multiple entangled geometric and photometric distortions (warps, perspective distortions, shadows, blurs, translations, rotations, scaling, and non-uniform illumination) making it highly suitable for establishing baselines for complex real-world scenarios. To address such entangled distortions, we propose a unified restoration architecture, ENet-GP, which can resolve both geometric and photometric degradations simultaneously within a single-network-single-training setup. ENet-GP surpasses the multiple inference passes of unified model in terms of OCR metrics (CER and ED) and average wall-clock time, while delivering commendable visual quality. ENet-GP also achieves competitive text metric results on single-distortion benchmarks. This makes our model robust, versatile, and scalable making it highly appealing for casually captured real-world document images.

%
%
\bibliographystyle{splncs04}
\bibliography{main}

@String(CVPR  = {IEEE Conf. Comput. Vis. Pattern Recog.})

@String(ICCV  = {Int. Conf. Comput. Vis.})

@String(BMVC  = {Brit. Mach. Vis. Conf.})

@String(AAAI  = {AAAI})

@String(ICIP  = {IEEE Int. Conf. Image Process.})

@String(ICPR  = {Int. Conf. Pattern Recog.})

@String(ICASSP=	{ICASSP})

@String(CVPR  = {CVPR})

@String(ICCV  = {ICCV})

@String(BMVC  =	{BMVC})

@String(ICIP  = {ICIP})

@String(ICPR  = {ICPR})

@misc{zhang2024docresgeneralistmodelunifying,
      title={DocRes: A Generalist Model Toward Unifying Document Image Restoration Tasks}, 
      author={Jiaxin Zhang and Dezhi Peng and Chongyu Liu and Peirong Zhang and Lianwen Jin},
      year={2024},
      eprint={2405.04408},
      archivePrefix={arXiv},
      primaryClass={cs.CV},
      url={https://arxiv.org/abs/2405.04408}, 
}

@inproceedings{SagnikKeICCV2019, 
Author = {Sagnik Das* and Ke Ma* and Zhixin Shu and Dimitris Samaras and Roy Shilkrot}, 
Booktitle = {Proceedings of International Conference on Computer Vision}, 
Title = {DewarpNet: Single-Image Document Unwarping With Stacked 3D and 2D Regression Networks}, 
Year = {2019}}

@misc{feng2022doctrdocumentimagetransformer,
      title={DocTr: Document Image Transformer for Geometric Unwarping and Illumination Correction}, 
      author={Hao Feng and Yuechen Wang and Wengang Zhou and Jiajun Deng and Houqiang Li},
      year={2022},
      eprint={2110.12942},
      archivePrefix={arXiv},
      primaryClass={cs.CV},
      url={https://arxiv.org/abs/2110.12942}, 
}

@InProceedings{Li_2023_ICCV,
    author    = {Li, Heng and Wu, Xiangping and Chen, Qingcai and Xiang, Qianjin},
    title     = {Foreground and Text-lines Aware Document Image Rectification},
    booktitle = {Proceedings of the IEEE/CVF International Conference on Computer Vision (ICCV)},
    month     = {October},
    year      = {2023},
    pages     = {19574-19583}
}

@misc{wang2022udocganunpaireddocumentillumination,
      title={UDoc-GAN: Unpaired Document Illumination Correction with Background Light Prior}, 
      author={Yonghui Wang and Wengang Zhou and Zhenbo Lu and Houqiang Li},
      year={2022},
      eprint={2210.08216},
      archivePrefix={arXiv},
      primaryClass={cs.CV},
      url={https://arxiv.org/abs/2210.08216}, 
}

@InProceedings{Zhang_2023_CVPR,
    author    = {Zhang, Ling and He, Yinghao and Zhang, Qing and Liu, Zheng and Zhang, Xiaolong and Xiao, Chunxia},
    title     = {Document Image Shadow Removal Guided by Color-Aware Background},
    booktitle = {Proceedings of the IEEE/CVF Conference on Computer Vision and Pattern Recognition (CVPR)},
    month     = {June},
    year      = {2023},
    pages     = {1818-1827}
}

@InProceedings{Ma_2018_CVPR,
author = {Ma, Ke and Shu, Zhixin and Bai, Xue and Wang, Jue and Samaras, Dimitris},
title = {DocUNet: Document Image Unwarping via a Stacked U-Net},
booktitle = {Proceedings of the IEEE Conference on Computer Vision and Pattern Recognition (CVPR)},
month = {June},
year = {2018}
}

@misc{zhao2025unidocdiffunifieddocumentrestoration,
      title={Uni-DocDiff: A Unified Document Restoration Model Based on Diffusion}, 
      author={Fangmin Zhao and Weichao Zeng and Zhenhang Li and Dongbao Yang and Binbin Li and Xiaojun Bi and Yu Zhou},
      year={2025},
      eprint={2508.04055},
      archivePrefix={arXiv},
      primaryClass={cs.CV},
      url={https://arxiv.org/abs/2508.04055}, 
}

@inproceedings{zamir2022restormer,
  title={Restormer: Efficient transformer for high-resolution image restoration},
  author={Zamir, Syed Waqas and Arora, Aditya and Khan, Salman and Hayat, Munawar and Khan, Fahad Shahbaz and Yang, Ming-Hsuan},
  booktitle={Proceedings of the IEEE/CVF Conference on Computer Vision and Pattern Recognition (CVPR)},
  pages={5728--5739},
  year={2022}
}

@inproceedings{jiang2022revisiting,
  title={Revisiting document image dewarping by grid regularization},
  author={Jiang, Xiangwei and Ma, Ke and Shu, Zhixin and Samaras, Dimitris},
  booktitle={Proceedings of the IEEE/CVF Conference on Computer Vision and Pattern Recognition (CVPR)},
  pages={5132--5141},
  year={2022}
}

@INPROCEEDINGS{9156786,
  author={Lin, Yun-Hsuan and Chen, Wen-Chin and Chuang, Yung-Yu},
  booktitle={2020 IEEE/CVF Conference on Computer Vision and Pattern Recognition (CVPR)}, 
  title={BEDSR-Net: A Deep Shadow Removal Network From a Single Document Image}, 
  year={2020},
  volume={},
  number={},
  pages={12902-12911},
  doi={10.1109/CVPR42600.2020.01292}}

@inproceedings{hertlein2019scannet,
  title={ScannerNet: A deep network for scanner-quality image enhancement},
  author={Hertlein, Felix and Naumann, Alexander and Philipp, Patrick},
  booktitle={BMVC},
  year={2019}
}

@inproceedings{das2019dewarpnet,
  title={DewarpNet: Single-image document unwarping with stacked 3D and 2D regression networks},
  author={Das, Sagnik and Ma, Ke and Shu, Zhixin and Samaras, Dimitris and Shilkrot, Roy},
  booktitle={Proceedings of the IEEE/CVF International Conference on Computer Vision},
  pages={131--140},
  year={2019}
}

@inproceedings{feng2021docmae,
  title={DocMAE: Document image rectification via self-supervised representation learning},
  author={Feng, Hao and Zhou, Wengang and Deng, Jiajun and Wang, Yuechen and Li, Houqiang},
  booktitle={arXiv preprint arXiv:2304.10341},
  year={2023}
}

@misc{tensmeyer2017documentimagebinarizationfully,
      title={Document Image Binarization with Fully Convolutional Neural Networks}, 
      author={Chris Tensmeyer and Tony Martinez},
      year={2017},
      eprint={1708.03276},
      archivePrefix={arXiv},
      primaryClass={cs.CV},
      url={https://arxiv.org/abs/1708.03276}, 
}

@INPROCEEDINGS{8578350,
  author={Kligler, Netanel and Katz, Sagi and Tal, Ayellet},
  booktitle={2018 IEEE/CVF Conference on Computer Vision and Pattern Recognition}, 
  title={Document Enhancement Using Visibility Detection}, 
  year={2018},
  volume={},
  number={},
  pages={2374-2382},
  doi={10.1109/CVPR.2018.00252}}

@article{calvo2019selectional,
  title={Selectional auto-encoder approach for document image binarization},
  author={Calvo-Zaragoza, Jorge and Gallego, Antonio-Javier},
  journal={Pattern Recognition},
  volume={86},
  pages={37--47},
  year={2019}
}

@article{souibgui2020docentr,
  title={DocEnTr: An end-to-end document image enhancement transformer},
  author={Souibgui, Mohamed Ali and Kessentini, Yousri and Fornes, Alicia and Llados, Josep},
  journal={arXiv preprint arXiv:2201.10252},
  year={2022}
}

@INPROCEEDINGS{9879292,
  author={Li, Boyun and Liu, Xiao and Hu, Peng and Wu, Zhongqin and Lv, Jiancheng and Peng, Xi},
  booktitle={2022 IEEE/CVF Conference on Computer Vision and Pattern Recognition (CVPR)}, 
  title={All-In-One Image Restoration for Unknown Corruption}, 
  year={2022},
  volume={},
  number={},
  pages={17431-17441},
  doi={10.1109/CVPR52688.2022.01693}}

@misc{wang2021uformergeneralushapedtransformer,
      title={Uformer: A General U-Shaped Transformer for Image Restoration}, 
      author={Zhendong Wang and Xiaodong Cun and Jianmin Bao and Wengang Zhou and Jianzhuang Liu and Houqiang Li},
      year={2021},
      eprint={2106.03106},
      archivePrefix={arXiv},
      primaryClass={cs.CV},
      url={https://arxiv.org/abs/2106.03106}, 
}

@misc{kim2022ocrfreedocumentunderstandingtransformer,
      title={OCR-free Document Understanding Transformer}, 
      author={Geewook Kim and Teakgyu Hong and Moonbin Yim and Jeongyeon Nam and Jinyoung Park and Jinyeong Yim and Wonseok Hwang and Sangdoo Yun and Dongyoon Han and Seunghyun Park},
      year={2022},
      eprint={2111.15664},
      archivePrefix={arXiv},
      primaryClass={cs.LG},
      url={https://arxiv.org/abs/2111.15664}, 
}

@article{blecher2023nougat,
  title={Nougat: Neural optical understanding for academic documents},
  author={Blecher, Lukas and Cucurull, Guillem and Scialom, Thomas and Stojnic, Robert},
  journal={arXiv preprint arXiv:2308.13418},
  year={2023}
}

@misc{davis2022endtoenddocumentrecognitionunderstanding,
      title={End-to-end Document Recognition and Understanding with Dessurt}, 
      author={Brian Davis and Bryan Morse and Bryan Price and Chris Tensmeyer and Curtis Wigington and Vlad Morariu},
      year={2022},
      eprint={2203.16618},
      archivePrefix={arXiv},
      primaryClass={cs.CV},
      url={https://arxiv.org/abs/2203.16618}, 
}

@INPROCEEDINGS{4376991,
  author={Smith, R.},
  booktitle={Ninth International Conference on Document Analysis and Recognition (ICDAR 2007)}, 
  title={An Overview of the Tesseract OCR Engine}, 
  year={2007},
  volume={2},
  number={},
  pages={629-633},
  doi={10.1109/ICDAR.2007.4376991}}

@InProceedings{Hertlein_2025_WACV,
    author    = {Hertlein, Felix and Naumann, Alexander and Sure-Vetter, York},
    title     = {DocMatcher: Document Image Dewarping via Structural and Textual Line Matching},
    booktitle = {Proceedings of the Winter Conference on Applications of Computer Vision (WACV)},
    month     = {February},
    year      = {2025},
    pages     = {5771-5780}
}

@inproceedings{Hertlein_ICCVW2023,
  author    = {Hertlein, Felix and Naumann, Alexander and Philipp, Patrick},
  title     = {Template-Guided Illumination Correction for Document Images with Imperfect Geometric Reconstruction},
  booktitle = {ICCV Workshops},
  year      = {2023}
}

@inproceedings{zhang2023docbinformer,
  author    = {Zhang, Jiaxin and Peng, Dezhi and Liu, Chongyu and Zhang, Peirong and Jin, Lianwen},
  title     = {DocBinFormer: A Two-Level Transformer for Document Binarization},
  booktitle = {arXiv preprint arXiv:2312.03568},
  year      = {2023}
}

@inproceedings{Rezanezhad_2024_Hybrid,
  author    = {Rezanezhad, Ali and Martínez, José},
  title     = {Hybrid CNN-Transformer Model for Historical Document Binarization},
  booktitle = {ICDAR},
  year      = {2024}
}

@misc{das2020intrinsicdecompositiondocumentimages,
      title={Intrinsic Decomposition of Document Images In-the-Wild}, 
      author={Sagnik Das and Hassan Ahmed Sial and Ke Ma and Ramon Baldrich and Maria Vanrell and Dimitris Samaras},
      year={2020},
      eprint={2011.14447},
      archivePrefix={arXiv},
      primaryClass={cs.CV},
      url={https://arxiv.org/abs/2011.14447}, 
}

@ARTICLE{10268585,
  author={Zhang, Jiaxin and Liang, Lingyu and Ding, Kai and Guo, Fengjun and Jin, Lianwen},
  journal={IEEE Transactions on Artificial Intelligence}, 
  title={Appearance Enhancement for Camera-Captured Document Images in the Wild}, 
  year={2024},
  volume={5},
  number={5},
  pages={2319-2330},
  doi={10.1109/TAI.2023.3321257}}

@misc{abuolaim2020defocusdeblurringusingdualpixel,
      title={Defocus Deblurring Using Dual-Pixel Data}, 
      author={Abdullah Abuolaim and Michael S. Brown},
      year={2020},
      eprint={2005.00305},
      archivePrefix={arXiv},
      primaryClass={eess.IV},
      url={https://arxiv.org/abs/2005.00305}, 
}

@misc{cho2021rethinkingcoarsetofineapproachsingle,
      title={Rethinking Coarse-to-Fine Approach in Single Image Deblurring}, 
      author={Sung-Jin Cho and Seo-Won Ji and Jun-Pyo Hong and Seung-Won Jung and Sung-Jea Ko},
      year={2021},
      eprint={2108.05054},
      archivePrefix={arXiv},
      primaryClass={cs.CV},
      url={https://arxiv.org/abs/2108.05054}, 
}

@misc{zamir2021multistageprogressiveimagerestoration,
      title={Multi-Stage Progressive Image Restoration}, 
      author={Syed Waqas Zamir and Aditya Arora and Salman Khan and Munawar Hayat and Fahad Shahbaz Khan and Ming-Hsuan Yang and Ling Shao},
      year={2021},
      eprint={2102.02808},
      archivePrefix={arXiv},
      primaryClass={cs.CV},
      url={https://arxiv.org/abs/2102.02808}, 
}

@misc{mao2016imagerestorationusingdeep,
      title={Image Restoration Using Very Deep Convolutional Encoder-Decoder Networks with Symmetric Skip Connections}, 
      author={Xiao-Jiao Mao and Chunhua Shen and Yu-Bin Yang},
      year={2016},
      eprint={1603.09056},
      archivePrefix={arXiv},
      primaryClass={cs.CV},
      url={https://arxiv.org/abs/1603.09056}, 
}

@misc{he2015deepresiduallearningimage,
      title={Deep Residual Learning for Image Recognition}, 
      author={Kaiming He and Xiangyu Zhang and Shaoqing Ren and Jian Sun},
      year={2015},
      eprint={1512.03385},
      archivePrefix={arXiv},
      primaryClass={cs.CV},
      url={https://arxiv.org/abs/1512.03385}, 
}

@INPROCEEDINGS{8546199,
  author={Zhao, Guoping and Liu, Jiajun and Jiang, Jiacheng and Guan, Hua and Wen, Ji-Rong},
  booktitle={2018 24th International Conference on Pattern Recognition (ICPR)}, 
  title={Skip-Connected Deep Convolutional Autoencoder for Restoration of Document Images}, 
  year={2018},
  volume={},
  number={},
  pages={2935-2940},
  doi={10.1109/ICPR.2018.8546199}}

@misc{liang2021swinirimagerestorationusing,
      title={SwinIR: Image Restoration Using Swin Transformer}, 
      author={Jingyun Liang and Jiezhang Cao and Guolei Sun and Kai Zhang and Luc Van Gool and Radu Timofte},
      year={2021},
      eprint={2108.10257},
      archivePrefix={arXiv},
      primaryClass={eess.IV},
      url={https://arxiv.org/abs/2108.10257}, 
}

@inproceedings{Verhoeven_2023, series={SA ’23},
   title={UVDoc: Neural Grid-based Document Unwarping},
   url={http://dx.doi.org/10.1145/3610548.3618174},
   DOI={10.1145/3610548.3618174},
   booktitle={SIGGRAPH Asia 2023 Conference Papers},
   publisher={ACM},
   author={Verhoeven, Floor and Magne, Tanguy and Sorkine-Hornung, Olga},
   year={2023},
   month=dec, pages={1–11},
   collection={SA ’23} }

@misc{zhou2024docdeshadowerfrequencyawaretransformerdocument,
      title={DocDeshadower: Frequency-Aware Transformer for Document Shadow Removal}, 
      author={Ziyang Zhou and Yingtie Lei and Xuhang Chen and Shenghong Luo and Wenjun Zhang and Chi-Man Pun and Zhen Wang},
      year={2024},
      eprint={2307.15318},
      archivePrefix={arXiv},
      primaryClass={cs.CV},
      url={https://arxiv.org/abs/2307.15318}, 
}

@misc{shi2016realtimesingleimagevideo,
      title={Real-Time Single Image and Video Super-Resolution Using an Efficient Sub-Pixel Convolutional Neural Network}, 
      author={Wenzhe Shi and Jose Caballero and Ferenc Huszár and Johannes Totz and Andrew P. Aitken and Rob Bishop and Daniel Rueckert and Zehan Wang},
      year={2016},
      eprint={1609.05158},
      archivePrefix={arXiv},
      primaryClass={cs.CV},
      url={https://arxiv.org/abs/1609.05158}, 
}

@inproceedings{inproceedings,
author = {Hradis, Michal and Kotera, Jan and Zemcík, Pavel and Sroubek, Filip},
year = {2015},
month = {09},
pages = {},
title = {Convolutional Neural Networks for Direct Text Deblurring},
doi = {10.5244/C.29.6}
}

@article{zhang2025docaligner,
  title={DocAligner: Automating the Annotation of Photographed Documents Through Real-virtual Alignment},
  author={Zhang, Jiaxin and Zhang, Peirong and Cheng, Huiyi and Chen, Xinhong and Xu, Haowei and Ding, Kai and Jin, Lianwen},
  journal={Pattern Recognition},
  pages={112191},
  year={2025},
  publisher={Elsevier}
}

@INPROCEEDINGS{9897217,
  author={Matsuo, Yuhi and Akimoto, Naofumi and Aoki, Yoshimitsu},
  booktitle={2022 IEEE International Conference on Image Processing (ICIP)}, 
  title={Document Shadow Removal with Foreground Detection Learning From Fully Synthetic Images}, 
  year={2022},
  volume={},
  number={},
  pages={1656-1660},
  doi={10.1109/ICIP46576.2022.9897217}}

@INPROCEEDINGS{8583809,
  author={Pratikakis, Ioannis and Zagori, Konstantinos and Kaddas, Panagiotis and Gatos, Basilis},
  booktitle={2018 16th International Conference on Frontiers in Handwriting Recognition (ICFHR)}, 
  title={ICFHR 2018 Competition on Handwritten Document Image Binarization (H-DIBCO 2018)}, 
  year={2018},
  volume={},
  number={},
  pages={489-493},
  doi={10.1109/ICFHR-2018.2018.00091}}

@inproceedings{feng2022geometric,
  title={Geometric representation learning for document image rectification},
  author={Feng, Hao and Zhou, Wengang and Deng, Jiajun and Wang, Yuechen and Li, Houqiang},
  booktitle={European Conference on Computer Vision},
  pages={475--492},
  year={2022}
}

@Article{s20236929,
AUTHOR = {Wang, Bingshu and Chen, C. L. Philip},
TITLE = {Local Water-Filling Algorithm for Shadow Detection and Removal of Document Images},
JOURNAL = {Sensors},
VOLUME = {20},
YEAR = {2020},
NUMBER = {23},
ARTICLE-NUMBER = {6929},
URL = {https://www.mdpi.com/1424-8220/20/23/6929},
PubMedID = {33291572},
ISSN = {1424-8220},
DOI = {10.3390/s20236929}
}

@INPROCEEDINGS{10447446,
  author={Tang, Hao and Guo, Junyuan and Wang, Teng and Yu, Yanwei and Wang, Chao},
  booktitle={ICASSP 2024 - 2024 IEEE International Conference on Acoustics, Speech and Signal Processing (ICASSP)}, 
  title={Efficient Joint Rectification of Photometric and Geometric Distortions in Document Images}, 
  year={2024},
  volume={},
  number={},
  pages={3690-3694},
  doi={10.1109/ICASSP48485.2024.10447446}}

@misc{jung2019waterfillingefficientalgorithmdigitized,
      title={Water-Filling: An Efficient Algorithm for Digitized Document Shadow Removal}, 
      author={Seungjun Jung and Muhammad Abul Hasan and Changick Kim},
      year={2019},
      eprint={1904.09763},
      archivePrefix={arXiv},
      primaryClass={cs.CV},
      url={https://arxiv.org/abs/1904.09763}, 
}

@misc{ruder2017overviewmultitasklearningdeep,
      title={An Overview of Multi-Task Learning in Deep Neural Networks}, 
      author={Sebastian Ruder},
      year={2017},
      eprint={1706.05098},
      archivePrefix={arXiv},
      primaryClass={cs.LG},
      url={https://arxiv.org/abs/1706.05098}, 
}

@INPROCEEDINGS{9706885,
  author={Hickson, Steven and Raveendran, Karthik and Essa, Irfan},
  booktitle={2022 IEEE/CVF Winter Conference on Applications of Computer Vision (WACV)}, 
  title={Sharing Decoders: Network Fission for Multi-task Pixel Prediction}, 
  year={2022},
  volume={},
  number={},
  pages={3655-3664},
  doi={10.1109/WACV51458.2022.00371}}

@misc{ronneberger2015unetconvolutionalnetworksbiomedical,
      title={U-Net: Convolutional Networks for Biomedical Image Segmentation}, 
      author={Olaf Ronneberger and Philipp Fischer and Thomas Brox},
      year={2015},
      eprint={1505.04597},
      archivePrefix={arXiv},
      primaryClass={cs.CV},
      url={https://arxiv.org/abs/1505.04597}, 
}

@INPROCEEDINGS{1292216,
  author={Wang, Z. and Simoncelli, E.P. and Bovik, A.C.},
  booktitle={The Thrity-Seventh Asilomar Conference on Signals, Systems \& Computers, 2003}, 
  title={Multiscale structural similarity for image quality assessment}, 
  year={2003},
  volume={2},
  number={},
  pages={1398-1402 Vol.2},
  doi={10.1109/ACSSC.2003.1292216}}

@misc{you2016multiviewrectificationfoldeddocuments,
      title={Multiview Rectification of Folded Documents}, 
      author={Shaodi You and Yasuyuki Matsushita and Sudipta Sinha and Yusuke Bou and Katsushi Ikeuchi},
      year={2016},
      eprint={1606.00166},
      archivePrefix={arXiv},
      primaryClass={cs.CV},
      url={https://arxiv.org/abs/1606.00166}, 
}

@inproceedings{10.1145/3528233.3530756,
author = {Ma, Ke and Das, Sagnik and Shu, Zhixin and Samaras, Dimitris},
title = {Learning From Documents in the Wild to Improve Document Unwarping},
year = {2022},
isbn = {9781450393379},
publisher = {Association for Computing Machinery},
address = {New York, NY, USA},
url = {https://doi.org/10.1145/3528233.3530756},
doi = {10.1145/3528233.3530756},
booktitle = {ACM SIGGRAPH 2022 Conference Proceedings},
articleno = {34},
numpages = {9},
location = {Vancouver, BC, Canada},
series = {SIGGRAPH '22}
}

@misc{yang2023docdiffdocumentenhancementresidual,
      title={DocDiff: Document Enhancement via Residual Diffusion Models}, 
      author={Zongyuan Yang and Baolin Liu and Yongping Xiong and Lan Yi and Guibin Wu and Xiaojun Tang and Ziqi Liu and Junjie Zhou and Xing Zhang},
      year={2023},
      eprint={2305.03892},
      archivePrefix={arXiv},
      primaryClass={cs.CV},
      url={https://arxiv.org/abs/2305.03892}, 
}

@misc{feng2022geometricrepresentationlearningdocument,
      title={Geometric Representation Learning for Document Image Rectification}, 
      author={Hao Feng and Wengang Zhou and Jiajun Deng and Yuechen Wang and Houqiang Li},
      year={2022},
      eprint={2210.08161},
      archivePrefix={arXiv},
      primaryClass={cs.CV},
      url={https://arxiv.org/abs/2210.08161}, 
}

@misc{li2024highresolutiondocumentshadowremoval,
      title={High-Resolution Document Shadow Removal via A Large-Scale Real-World Dataset and A Frequency-Aware Shadow Erasing Net}, 
      author={Zinuo Li and Xuhang Chen and Chi-Man Pun and Xiaodong Cun},
      year={2024},
      eprint={2308.14221},
      archivePrefix={arXiv},
      primaryClass={cs.CV},
      url={https://arxiv.org/abs/2308.14221}, 
}

@inproceedings{wang2024docnlc,
  title={DocNLC: A Document Image Enhancement Framework with Normalized and Latent Contrastive Representation for Multiple Degradations},
  author={Wang, Ruilu and Xue, Yang and Jin, Lianwen},
  booktitle={Proceedings of the AAAI Conference on Artificial Intelligence},
  volume={38},
  number={6},
  pages={5563--5571},
  year={2024}
}

@misc{ma2023proresexploringdegradationawarevisual,
      title={ProRes: Exploring Degradation-aware Visual Prompt for Universal Image Restoration}, 
      author={Jiaqi Ma and Tianheng Cheng and Guoli Wang and Qian Zhang and Xinggang Wang and Lefei Zhang},
      year={2023},
      eprint={2306.13653},
      archivePrefix={arXiv},
      primaryClass={cs.CV},
      url={https://arxiv.org/abs/2306.13653}, 
}

@INPROCEEDINGS{5277767,
  author={Gatos, Basilis and Ntirogiannis, Konstantinos and Pratikakis, Ioannis},
  booktitle={2009 10th International Conference on Document Analysis and Recognition}, 
  title={ICDAR 2009 Document Image Binarization Contest (DIBCO 2009)}, 
  year={2009},
  volume={},
  number={},
  pages={1375-1382},
  doi={10.1109/ICDAR.2009.246}}

@INPROCEEDINGS{6981120,
  author={Ntirogiannis, Konstantinos and Gatos, Basilis and Pratikakis, Ioannis},
  booktitle={2014 14th International Conference on Frontiers in Handwriting Recognition}, 
  title={ICFHR2014 Competition on Handwritten Document Image Binarization (H-DIBCO 2014)}, 
  year={2014},
  volume={},
  number={},
  pages={809-813},
  doi={10.1109/ICFHR.2014.141}}

@INPROCEEDINGS{5693650,
  author={Pratikakis, Ioannis and Gatos, Basilis and Ntirogiannis, Konstantinos},
  booktitle={2010 12th International Conference on Frontiers in Handwriting Recognition}, 
  title={H-DIBCO 2010 - Handwritten Document Image Binarization Competition}, 
  year={2010},
  volume={},
  number={},
  pages={727-732},
  doi={10.1109/ICFHR.2010.118}}

@INPROCEEDINGS{6065249,
  author={Pratikakis, Ioannis and Gatos, Basilis and Ntirogiannis, Konstantinos},
  booktitle={2011 International Conference on Document Analysis and Recognition}, 
  title={ICDAR 2011 Document Image Binarization Contest (DIBCO 2011)}, 
  year={2011},
  volume={},
  number={},
  pages={1506-1510},
  doi={10.1109/ICDAR.2011.299}}

@INPROCEEDINGS{6424498,
  author={Pratikakis, Ioannis and Gatos, Basilis and Ntirogiannis, Konstantinos},
  booktitle={2012 International Conference on Frontiers in Handwriting Recognition}, 
  title={ICFHR 2012 Competition on Handwritten Document Image Binarization (H-DIBCO 2012)}, 
  year={2012},
  volume={},
  number={},
  pages={817-822},
  doi={10.1109/ICFHR.2012.216}}

@INPROCEEDINGS{6628857,
  author={Pratikakis, Ioannis and Gatos, Basilis and Ntirogiannis, Konstantinos},
  booktitle={2013 12th International Conference on Document Analysis and Recognition}, 
  title={ICDAR 2013 Document Image Binarization Contest (DIBCO 2013)}, 
  year={2013},
  volume={},
  number={},
  pages={1471-1476},
  doi={10.1109/ICDAR.2013.219}}

@INPROCEEDINGS{7814134,
  author={Pratikakis, Ioannis and Zagoris, Konstantinos and Barlas, George and Gatos, Basilis},
  booktitle={2016 15th International Conference on Frontiers in Handwriting Recognition (ICFHR)}, 
  title={ICFHR2016 Handwritten Document Image Binarization Contest (H-DIBCO 2016)}, 
  year={2016},
  volume={},
  number={},
  pages={619-623},
  doi={10.1109/ICFHR.2016.0118}}

@INPROCEEDINGS{8270159,
  author={Pratikakis, Ioannis and Zagoris, Konstantinos and Barlas, George and Gatos, Basilis},
  booktitle={2017 14th IAPR International Conference on Document Analysis and Recognition (ICDAR)}, 
  title={ICDAR2017 Competition on Document Image Binarization (DIBCO 2017)}, 
  year={2017},
  volume={01},
  number={},
  pages={1395-1403},
  doi={10.1109/ICDAR.2017.228}}

@INPROCEEDINGS{8978205,
  author={Pratikakis, Ioannis and Zagoris, Konstantinos and Karagiannis, Xenofon and Tsochatzidis, Lazaros and Mondal, Tanmoy and Marthot-Santaniello, Isabelle},
  booktitle={2019 International Conference on Document Analysis and Recognition (ICDAR)}, 
  title={ICDAR 2019 Competition on Document Image Binarization (DIBCO 2019)}, 
  year={2019},
  volume={},
  number={},
  pages={1547-1556},
  doi={10.1109/ICDAR.2019.00249}}

@inproceedings{10.5555/1768409.1768429,
author = {Zamora-Mart\'{\i}nez, F. and Espa\~{n}a-Boquera, S. and Castro-Bleda, M. J.},
title = {Behaviour-based clustering of neural networks applied to document enhancement},
year = {2007},
isbn = {9783540730064},
publisher = {Springer-Verlag},
address = {Berlin, Heidelberg},
booktitle = {Proceedings of the 9th International Work Conference on Artificial Neural Networks},
pages = {144–151},
numpages = {8},
location = {San Sebasti\'{a}n, Spain},
series = {IWANN'07}
}

@INPROCEEDINGS{7333947,
  author={Hedjam, Rachid and Nafchi, Hossein Ziaei and Moghaddam, Reza Farrahi and Kalacska, Margaret and Cheriet, Mohamed},
  booktitle={2015 13th International Conference on Document Analysis and Recognition (ICDAR)}, 
  title={ICDAR 2015 contest on MultiSpectral Text Extraction (MS-TEx 2015)}, 
  year={2015},
  volume={},
  number={},
  pages={1181-1185},
  doi={10.1109/ICDAR.2015.7333947}}

@inproceedings{Ayatollahi_2013,
   title={Persian heritage image binarization competition (PHIBC 2012)},
   url={http://dx.doi.org/10.1109/PRIA.2013.6528442},
   DOI={10.1109/pria.2013.6528442},
   booktitle={2013 First Iranian Conference on Pattern Recognition and Image Analysis (PRIA)},
   publisher={IEEE},
   author={Ayatollahi, S. M. and Nafchi, H. Z.},
   year={2013},
   month=mar, pages={1–4} }

@inproceedings{10.1145/1816123.1816161,
author = {Deng, Fanbo and Wu, Zheng and Lu, Zheng and Brown, Michael S.},
title = {BinarizationShop: a user-assisted software suite for converting old documents to black-and-white},
year = {2010},
isbn = {9781450300858},
publisher = {Association for Computing Machinery},
address = {New York, NY, USA},
url = {https://doi.org/10.1145/1816123.1816161},
doi = {10.1145/1816123.1816161},
booktitle = {Proceedings of the 10th Annual Joint Conference on Digital Libraries},
pages = {255–258},
numpages = {4},
location = {Gold Coast, Queensland, Australia},
series = {JCDL '10}
}

@InProceedings{Sayed_2021_CVPR,
  author    = {Sayed, Mohamed and Brostow, Gabriel},
  title     = {Improved Handling of Motion Blur in Online Object Detection},
  booktitle = {Proceedings of the IEEE/CVF Conference on Computer Vision and Pattern Recognition (CVPR)},
  month     = {June},
  year      = {2021},
  pages     = {1706-1716}
}

@misc{zhang2025dvdunleashinggenerativeparadigm,
      title={DvD: Unleashing a Generative Paradigm for Document Dewarping via Coordinates-based Diffusion Model}, 
      author={Weiguang Zhang and Huangcheng Lu and Maizhen Ning and Xiaowei Huang and Wei Wang and Kaizhu Huang and Qiufeng Wang},
      year={2025},
      eprint={2505.21975},
      archivePrefix={arXiv},
      primaryClass={cs.CV},
      url={https://arxiv.org/abs/2505.21975}, 
}

@InProceedings{hertlein2025docmatcher,
    author    = {Hertlein, Felix and Naumann, Alexander and Sure-Vetter, York},
    title     = {DocMatcher: Document Image Dewarping via Structural and Textual Line Matching},
    booktitle = {Proceedings of the Winter Conference on Applications of Computer Vision (WACV)},
    month     = {February},
    year      = {2025},
    pages     = {5771-5780}
}

@misc{feng2022docscannerrobustdocumentimage,
      title={DocScanner: Robust Document Image Rectification with Progressive Learning}, 
      author={Hao Feng and Wengang Zhou and Jiajun Deng and Qi Tian and Houqiang Li},
      year={2022},
      eprint={2110.14968},
      archivePrefix={arXiv},
      primaryClass={cs.CV},
      url={https://arxiv.org/abs/2110.14968}, 
}

@InProceedings{Yu_2024_WACV,
    author    = {Yu, Fangchen and Xie, Yina and Wu, Lei and Wen, Yafei and Wang, Guozhi and Ren, Shuai and Chen, Xiaoxin and Mao, Jianfeng and Li, Wenye},
    title     = {DocReal: Robust Document Dewarping of Real-Life Images via Attention-Enhanced Control Point Prediction},
    booktitle = {Proceedings of the IEEE/CVF Winter Conference on Applications of Computer Vision (WACV)},
    month     = {January},
    year      = {2024},
    pages     = {665-674}
}

@INPROCEEDINGS{9880145,
  author={Xue, Chuhui and Tian, Zichen and Zhan, Fangneng and Lu, Shijian and Bai, Song},
  booktitle={2022 IEEE/CVF Conference on Computer Vision and Pattern Recognition (CVPR)}, 
  title={Fourier Document Restoration for Robust Document Dewarping and Recognition}, 
  year={2022},
  volume={},
  number={},
  pages={4563-4572},
  doi={10.1109/CVPR52688.2022.00453}
}

@article{Hertlein2023,
  title        = {Inv3D: a high-resolution 3D invoice dataset for template-guided single-image document unwarping},
  author       = {Hertlein, Felix and Naumann, Alexander and Philipp, Patrick},
  year         = 2023,
  month        = {Apr},
  day          = 29,
  journal      = {International Journal on Document Analysis and Recognition (IJDAR)},
  doi          = {10.1007/s10032-023-00434-x},
  ISSN         = {1433-2825},
  url          = {https://doi.org/10.1007/s10032-023-00434-x}
}

@misc{xie2022documentdewarpingcontrolpoints,
      title={Document Dewarping with Control Points}, 
      author={Guo-Wang Xie and Fei Yin and Xu-Yao Zhang and Cheng-Lin Liu},
      year={2022},
      eprint={2203.10543},
      archivePrefix={arXiv},
      primaryClass={cs.CV},
      url={https://arxiv.org/abs/2203.10543}, 
}

@article{10.1145/3627818,
author = {Li, Pu and Quan, Weize and Guo, Jianwei and Yan, Dong-Ming},
title = {Layout-aware Single-image Document Flattening},
year = {2023},
issue_date = {February 2024},
publisher = {Association for Computing Machinery},
address = {New York, NY, USA},
volume = {43},
number = {1},
issn = {0730-0301},
url = {https://doi.org/10.1145/3627818},
doi = {10.1145/3627818},
journal = {ACM Trans. Graph.},
month = nov,
articleno = {9},
numpages = {17}
}

@inproceedings{10.1145/3664647.3681548,
author = {Zhang, Weiguang and Wang, Qiufeng and Huang, Kaizhu and Huang, Xiaowei and Guo, Fengjun and Gu, Xiaomeng},
title = {Document Registration: Towards Automated Labeling of Pixel-Level Alignment Between Warped-Flat Documents},
year = {2024},
isbn = {9798400706868},
publisher = {Association for Computing Machinery},
address = {New York, NY, USA},
url = {https://doi.org/10.1145/3664647.3681548},
doi = {10.1145/3664647.3681548},
booktitle = {Proceedings of the 32nd ACM International Conference on Multimedia},
pages = {9933–9942},
numpages = {10},
location = {Melbourne VIC, Australia},
series = {MM '24}
}

@ARTICLE{10374269,
  author={Feng, Hao and Liu, Shaokai and Deng, Jiajun and Zhou, Wengang and Li, Houqiang},
  journal={IEEE Transactions on Multimedia}, 
  title={Deep Unrestricted Document Image Rectification}, 
  year={2024},
  volume={26},
  number={},
  pages={6142-6154},
  doi={10.1109/TMM.2023.3347094}}

@misc{liu2026booknetbookimagerectification,
      title={BookNet: Book Image Rectification via Cross-Page Attention Network}, 
      author={Shaokai Liu and Hao Feng and Bozhi Luan and Min Hou and Jiajun Deng and Wengang Zhou},
      year={2026},
      eprint={2601.21938},
      archivePrefix={arXiv},
      primaryClass={cs.CV},
      url={https://arxiv.org/abs/2601.21938}, 
}

@article{article3,
author = {Michalak, Hubert and Okarma, Krzysztof},
year = {2020},
month = {05},
pages = {},
title = {Robust Combined Binarization Method of Non-Uniformly Illuminated Document Images for Alphanumerical Character Recognition},
volume = {20},
journal = {Sensors},
doi = {10.3390/s20102914}
}

@article{article2,
author = {Matsuo, Yuhi and Aoki, Yoshimitsu},
year = {2024},
month = {01},
pages = {654},
title = {Synthetic Document Images with Diverse Shadows for Deep Shadow Removal Networks},
volume = {24},
journal = {Sensors},
doi = {10.3390/s24020654}
}

@article{10.1145/3355089.3356563,
author = {Li, Xiaoyu and Zhang, Bo and Liao, Jing and Sander, Pedro V.},
title = {Document rectification and illumination correction using a patch-based CNN},
year = {2019},
issue_date = {December 2019},
publisher = {Association for Computing Machinery},
address = {New York, NY, USA},
volume = {38},
number = {6},
issn = {0730-0301},
url = {https://doi.org/10.1145/3355089.3356563},
doi = {10.1145/3355089.3356563},
journal = {ACM Trans. Graph.},
month = nov,
articleno = {168},
numpages = {11}
}

@misc{xie2021dewarpingdocumentimagedisplacement,
      title={Dewarping Document Image By Displacement Flow Estimation with Fully Convolutional Network}, 
      author={Guo-Wang Xie and Fei Yin and Xu-Yao Zhang and Cheng-Lin Liu},
      year={2021},
      eprint={2104.06815},
      archivePrefix={arXiv},
      primaryClass={cs.CV},
      url={https://arxiv.org/abs/2104.06815}, 
}

@INPROCEEDINGS{9710331,
  author={Das, Sagnik and Singh, Kunwar Yashraj and Wu, Jon and Bas, Erhan and Mahadevan, Vijay and Bhotika, Rahul and Samaras, Dimitris},
  booktitle={2021 IEEE/CVF International Conference on Computer Vision (ICCV)}, 
  title={End-to-end Piece-wise Unwarping of Document Images}, 
  year={2021},
  volume={},
  number={},
  pages={4248-4257},
  doi={10.1109/ICCV48922.2021.00423}}

@Article{s18041135,
AUTHOR = {Li, Jinyang and Liu, Zhijing and Yao, Yong},
TITLE = {Defocus Blur Detection and Estimation from Imaging Sensors},
JOURNAL = {Sensors},
VOLUME = {18},
YEAR = {2018},
NUMBER = {4},
ARTICLE-NUMBER = {1135},
URL = {https://www.mdpi.com/1424-8220/18/4/1135},
PubMedID = {29642491},
ISSN = {1424-8220},
DOI = {10.3390/s18041135}
}

@Article{s150100880,
AUTHOR = {Cheong, Hejin and Chae, Eunjung and Lee, Eunsung and Jo, Gwanghyun and Paik, Joonki},
TITLE = {Fast Image Restoration for Spatially Varying Defocus Blur of Imaging Sensor},
JOURNAL = {Sensors},
VOLUME = {15},
YEAR = {2015},
NUMBER = {1},
PAGES = {880--898},
URL = {https://www.mdpi.com/1424-8220/15/1/880},
PubMedID = {25569760},
ISSN = {1424-8220},
DOI = {10.3390/s150100880}
}

@misc{liu2018intriguingfailingconvolutionalneural,
      title={An Intriguing Failing of Convolutional Neural Networks and the CoordConv Solution}, 
      author={Rosanne Liu and Joel Lehman and Piero Molino and Felipe Petroski Such and Eric Frank and Alex Sergeev and Jason Yosinski},
      year={2018},
      eprint={1807.03247},
      archivePrefix={arXiv},
      primaryClass={cs.CV},
      url={https://arxiv.org/abs/1807.03247}, 
}

@INPROCEEDINGS{6909892,
  author={Meng, Gaofeng and Wang, Ying and Qu, Shenquan and Xiang, Shiming and Pan, Chunhong},
  booktitle={2014 IEEE Conference on Computer Vision and Pattern Recognition}, 
  title={Active Flattening of Curved Document Images via Two Structured Beams}, 
  year={2014},
  volume={},
  number={},
  pages={3890-3897},
  doi={10.1109/CVPR.2014.497}}

@misc{istomin2025efficientdocumentimagedewarping,
      title={Efficient Document Image Dewarping via Hybrid Deep Learning and Cubic Polynomial Geometry Restoration}, 
      author={Valery Istomin and Oleg Pereziabov and Ilya Afanasyev},
      year={2025},
      eprint={2501.03145},
      archivePrefix={arXiv},
      primaryClass={cs.CV},
      url={https://arxiv.org/abs/2501.03145}, 
}
\end{document}